\documentclass[letterpaper,10pt,conference]{ieeeconf}

\IEEEoverridecommandlockouts %
\PassOptionsToPackage{table}{xcolor} %
\usepackage{amsmath} %
\usepackage{amssymb} %
\usepackage{amsfonts}
\usepackage[sort,compress]{cite}
\usepackage[usenames,dvipsnames]{xcolor}
\usepackage{graphicx}
\usepackage{tabularx}
\usepackage{multirow}
\usepackage{mathtools}
\usepackage{esvect} %
\usepackage[]{siunitx}
\usepackage{caption}
\usepackage[nolist]{acronym}
\usepackage[pdftex,pdfstartview={FitV},pdfpagelayout={TwoColumnLeft},bookmarksopen=true,plainpages=false,colorlinks=true,linkcolor=black,citecolor=black,urlcolor=black,filecolor=black ,pagebackref=false,hypertexnames=false,plainpages=false,pdfpagelabels]{hyperref}
\usepackage[T1]{fontenc} %
\usepackage{mathtools, cuted} %
\usepackage{bm}
\usepackage{xspace}
\usepackage{arydshln}

\usepackage{etoolbox}
\usepackage[normalem]{ulem}
\usepackage{empheq}
\usepackage{cleveref}
\robustify\bfseries
\robustify\uline

\newcolumntype{H}{>{\setbox0=\hbox\bgroup}c<{\egroup}@{}}

\usepackage{enumerate}
\usepackage{balance} %
\usepackage{booktabs} %

\newcommand{\VeryEasy}{\textbf{\textcolor{LimeGreen}{Very Easy}}}
\newcommand{\Easy}{\textbf{\textcolor{ForestGreen}{Easy}}}
\newcommand{\Moderate}{\textbf{\textcolor{YellowOrange}{Moderate}}}
\newcommand{\Hard}{\textbf{\textcolor{RedOrange}{Hard}}}
\newcommand{\VeryHard}{\textbf{\textcolor{Red}{Very Hard}}}

\newcommand{\intjerksquared}{\ensuremath{\int_{t_{in}}^{T}\left\Vert\mathbf{j}\right\Vert^2dt}}
\newcommand{\intddyawsquared}{\ensuremath{\int_{t_{in}}^{T}(\ddot{\psi})^2dt}}

\usepackage{makecell}

\usepackage{amsmath,calc}

\usepackage{tikz} %
\usetikzlibrary{shapes,snakes} %
\usetikzlibrary{tikzmark} %
\usetikzlibrary{calc, arrows.meta, positioning} %
\usetikzlibrary{shapes.multipart} %
\usepackage{subcaption} %

\usepackage{footnote}
\makesavenoteenv{tabular}
\makesavenoteenv{table}

\ifdefined\qtyproduct
\else
  \ifdefined\NewCommandCopy
    \NewCommandCopy\qtyproduct\SI
  \else
    \NewDocumentCommand\qtyproduct{O{}mm}{\SI[#1]{#2}{#3}}
  \fi
\fi

\DeclareCaptionLabelSeparator{periodspace}{.\quad}
\definecolor{opt_color}{RGB}{203,255,182}
\definecolor{check_color}{RGB}{195,218,255}
\definecolor{recheck_color}{RGB}{255,176,176}
\definecolor{delaycheck_color}{RGB}{255,228,181}

\definecolor{agent1_color}{RGB}{234,153,153}
\definecolor{agent2_color}{RGB}{151,104,175}
\definecolor{agent3_color}{RGB}{249,203,156}
\definecolor{agent4_color}{RGB}{194,115,156}
\definecolor{agent5_color}{RGB}{159,197,232}
\definecolor{agent6_color}{RGB}{117,186,117}
\definecolor{obstacle_color}{RGB}{160,117,109}

\usepackage{mdframed}

\newacro{slam}[SLAM]{Simultaneous Localization and Mapping}
\newacro{uav}[UAV]{Unmanned Aerial Vehicle}
\acrodefplural{uav}[UAVs]{unmanned aerial vehicles}
\newacro{gns}[GNS]{Global Navigation Satellite}
\newacro{gnss}[GNSS]{Global Navigation Satellite System}
\acrodefplural{gnss}[GNSS]{Global Navigation Satellite Systems}
\newacro{mcl}[MCL]{Monte-Carlo localization}
\newacro{imu}[IMU]{Inertial Measurement Unit}
\newacro{dof}[DOF]{degree-of-freedom}
\acrodefplural{dof}[DOFs]{degrees-of-freedom}
\newacro{ransac}[RANSAC]{random sample consensus}
\newacro{map}[MAP]{maximum a posteriori}
\newacro{mle}[MLE]{maximum likelihood estimation}
\newacro{rms}[RMS]{root-mean-square}
\newacro{dem}[DEM]{digital elevation model}
\newacro{vio}[VIO]{visual-inertial odometry}
\newacro{cnn}[CNN]{convolutional neural network}
\newacro{pdf}[pdf]{probability density function}
\newacro{ahrs}[AHRS]{attitude and heading reference system}
\newacro{lidar}[LIDAR]{light detection and ranging}
\newacro{relu}[ReLU]{rectified linear unit}
\newacro{rtk}[RTK]{real-time kinematic}
\newacro{gps}[GPS]{global positioning system}
\newacro{fcn}[FCN]{fully-connected network}
\newacro{brm}[BRM]{building ratio map}
\newacro{sfm}[SfM]{Structure-from-Motion}
\newacro{vpr}[VPR]{visual place recognition}
\newacro{fov}[FOV]{field of view}
\newacro{poc}[POC]{partially overlapping circular}

\title{\LARGE \bf PUMA: Fully Decentralized Uncertainty-aware Multiagent Trajectory Planner with Real-time Image Segmentation-based Frame Alignment}

\author{Kota Kondo, Claudius T.\ Tewari, Mason B. Peterson, Annika Thomas, \\ Jouko Kinnari, Andrea Tagliabue,  Jonathan P.\ How%
	\thanks{The authors are with the Department of Aeronautics and Astronautics, Massachusetts Institute of Technology.
	    {\texttt{\{kkondo, cttewari, masonbp, annikat, jkinnari, atagliab, jhow\}@mit.edu.}}}
    \thanks{This work is supported by Boeing Research \& Technology and the Air Force Office of Scientific Research MURI FA9550-19-1-0386.}
}%

\begin{document}

\maketitle
\thispagestyle{plain}
\pagestyle{plain}

\begin{abstract} 
Fully decentralized, multiagent trajectory planners enable complex tasks like search and rescue or package delivery by ensuring safe navigation in unknown environments.
However, deconflicting trajectories with other agents and ensuring collision-free paths in a fully decentralized setting is complicated by dynamic elements and localization uncertainty.
To this end, this paper presents (1) an uncertainty-aware multiagent trajectory planner and (2) an image segmentation-based frame alignment pipeline.
The uncertainty-aware planner propagates uncertainty associated with the future motion of detected obstacles, and by incorporating this propagated uncertainty into optimization constraints, the planner effectively navigates around obstacles. 
Unlike conventional methods that emphasize explicit obstacle tracking, our approach integrates implicit tracking. 
Moreover, sharing trajectories between agents can cause potential collisions due to frame misalignment.
Addressing this, we introduce a novel frame alignment pipeline that rectifies inter-agent frame misalignment. 
This method leverages a zero-shot image segmentation model for detecting objects in the environment and a data association framework based on geometric consistency for map alignment.
Our approach accurately aligns frames with only 0.18 m and 2.7$^\mathbf{\circ}$ of mean frame alignment error in our most challenging simulation scenario.
In addition, we conducted hardware experiments and successfully achieved 0.29 m and 2.59$^\mathbf{\circ}$ of frame alignment error.
Together with the alignment framework, our planner ensures safe navigation in unknown environments and collision avoidance in decentralized settings.
\acresetall
\end{abstract}

\section*{Supplementary Material}%
\noindent\textbf{Video}: \href{https://youtu.be/W73p42XRcaQ}{https://youtu.be/W73p42XRcaQ} \\
\textbf{Code}: \href{https://github.com/mit-acl/puma}{https://github.com/mit-acl/puma}

\section{Introduction}\label{sec:introduction}

\begin{figure}[t]
    \centering
    \subcaptionbox{Multiagent Frame Alignment Evaluation Hardware  Environment. \label{fig:hw-multi-agent-frame-alignment-env}}{
    \begin{tikzpicture}[every text node part/.style={align=center}]
        \node {\includegraphics[width=\columnwidth, trim={0 0 0 4cm}, clip]{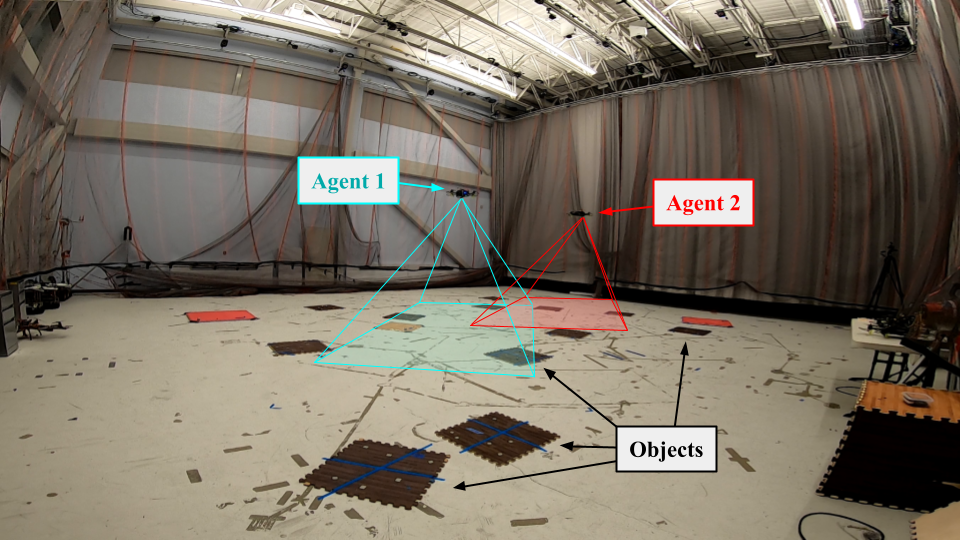}};
    \end{tikzpicture}}
    \subcaptionbox{Tracking Quality in Case 30 (difficulty: \Moderate): Our pipeline successfully estimates the drifted state of vehicle 1 even in multiagent environments. The estimated trajectory demonstrates low errors, indicated in green, and closely aligns with the ground truth trajectory. The estimated poses also exhibit a high degree of overlap with the ground truth, as shown by the significant overlapping of circles. The coordinate frames are captured at intervals of \SI{5}{\s}.\label{fig:multiagent-pcc-tracking-quality}}{
    \begin{tikzpicture}[every text node part/.style={align=center}]
        \node {\includegraphics[clip, trim=3.8in 2in 4.5in 1in, width=\columnwidth]{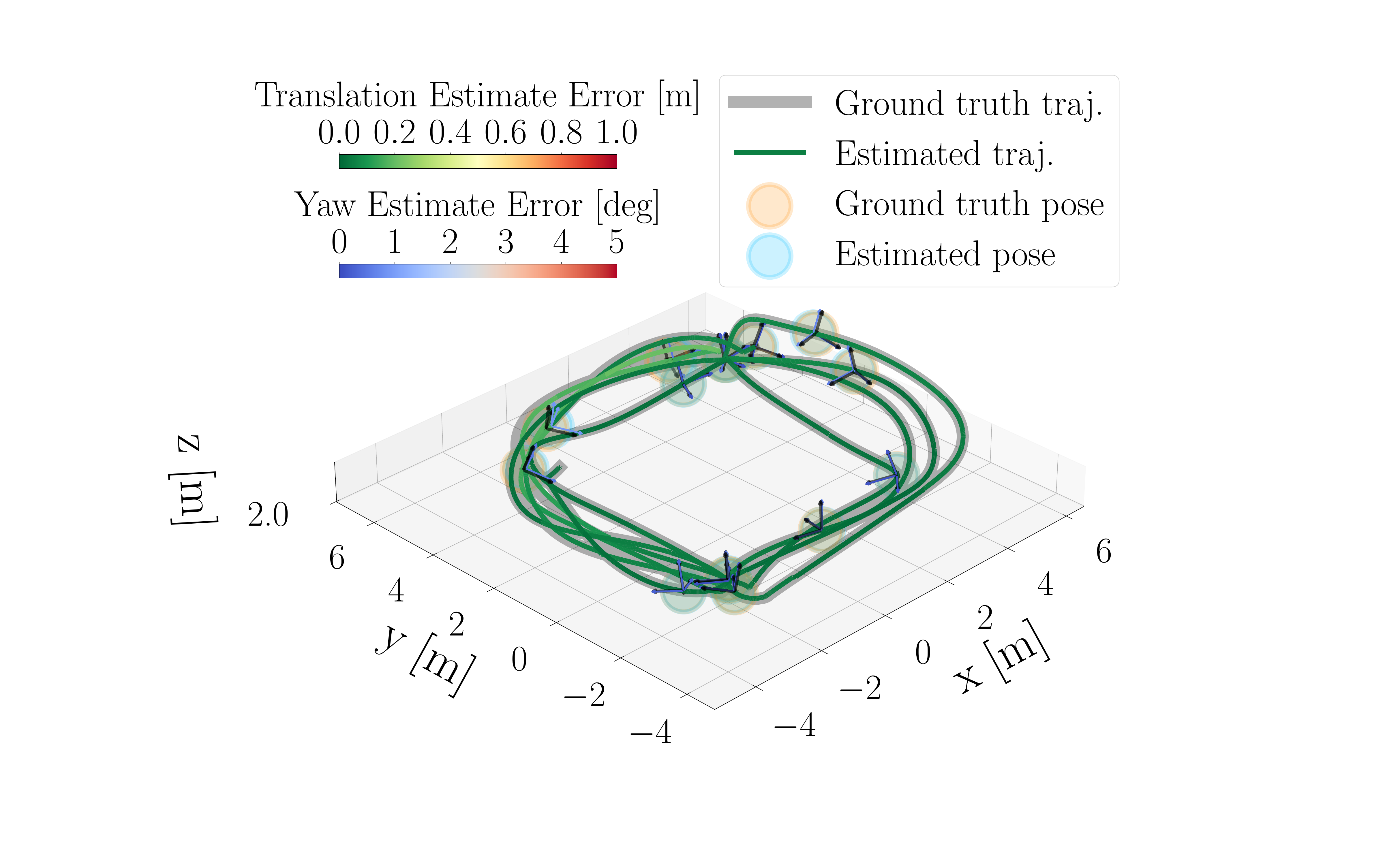}};
    \end{tikzpicture}}
    \caption{\textbf{PUMA} is thoroughly evaluated in simulation and hardware experiments are also performed to evaluate the real-time image segmentation-based frame alignment pipeline.} 
    \vspace*{-14pt}
\end{figure}

 Multiagent \ac{uav} trajectory planning has been extensively studied~\cite{ryou_cooperative_2022, peng2022obstacle, gao2022meeting, tordesillas2020mader, kondo2023robust, zhou2020ego-swarm, sebetghadam2022distributed, robinson2018efficient, park2020efficient, Hou2022EnhancedDA, firoozi2020distributed, toumieh2023decentralized, wang2022robust, batra2022decentralized, csenbacslar2023dream, csenbacslar2023mrnav}.
In real-world deployments of multiagent trajectory planning methods, it is crucial to (1) detect and avoid collisions with \textbf{obstacles}, (2)~correct for \textbf{localization errors and uncertainties}, and (3)~achieve \textbf{scalability} to a large number of agents. 

Equipping agents with sensors, often cameras, is a common method to detect and avoid unfamiliar obstacles~\cite{thomas2017autonomous,penin2017vision-based, falanga2018pampc,murali2019perception,spasojevic2020perception,tordesillas2021panther,zhou2021raptor,wu2022perception}.
This provides agents with real-time situational awareness, facilitating informed decisions for collision avoidance in dynamic settings. However, these sensors typically have a restricted \ac{fov}, and therefore the orientation of the \ac{uav} must be considered when planning trajectories through new spaces, and hence planners with limited FOV must be perception-aware to reduce environment uncertainty.

In addition, while many existing planners operate under the assumption of perfect global localization, often relying on \acp{gnss}, for practical real-world applications where perfect global localization is not available, agents need to perform onboard localization, such as \ac{vio}.
However, onboard localization in environments without global information introduces the significant challenge of overcoming state estimation drift. 
Because estimation drift corrupts an agent's estimate of its own location, any other spatial information that is considered relative to the agent's location (e.g. future trajectories and locations of observed obstacles) will be corrupted as well. 
So for a decentralized planning pipeline to function, it is important to both estimate the true relative poses between agents and to understand the uncertainty associated with that estimate. 
If these quantities are accurately represented, agents can effectively communicate about their trajectories and environment.

In multiagent trajectory planning, there are two approaches: centralized and decentralized. 
Centralized planners~\cite{robinson2018efficient, park2020efficient} have one entity that directs all agent trajectories, providing simplicity but with the drawback of a potential single point of failure.
Conversely, decentralized planners~\cite{zhou2020ego-swarm, tordesillas2020mader, sebetghadam2022distributed, kondo2023robust, toumieh2023decentralized, batra2022decentralized} enable each agent to plan its own trajectory, offering better scalability and resilience.
Similarly synchronous planners~\cite{van2017distributed, sebetghadam2022distributed, firoozi2020distributed} require synchronization before planning, whereas asynchronous methods~\cite{zhou2020ego-swarm, kondo2023robust, tordesillas2020mader} let agents plan individually, proving to be more scalable.

To tackle the aforementioned challenges: (1)~\textbf{unknown object detection and collision avoidance}, (2)~\textbf{localization errors and uncertainties}, and (3)~\textbf{scalability}, we introduce \textbf{PUMA}. Our contributions include:
\begin{enumerate}
    \item A fully decentralized, asynchronous, \textbf{P}erception-aware, and \textbf{U}ncertainty-aware \textbf{M}ulti\textbf{A}gent trajectory planner. 
    \item An inter-agent frame alignment pipeline that utilizes an image segmentation technique and a data association approach using geometric consistency.
    \item Comprehensive simulation benchmarks, comparing PUMA with a state-of-the-art perception-aware planner in \textbf{100} simulations, and thorough evaluation of our segmentation-based frame alignment pipeline across \textbf{42} different scenarios.
    \item Hardware tests on \ac{uav}s, demonstrating real-time onboard frame alignment, ensuring successful frame synchronization.
\end{enumerate}
\section{Uncertainty-Aware Multiagent Trajectory Planning}\label{sec:ua-trajectory}

This section describes our uncertainty-aware planner. 
Our planner generates collision-free position and yaw trajectories in a decentralized manner while simultaneously accounting for various uncertainties in the environment. 
Specifically, it tackles uncertainties from (1) the prediction of previously detected objects and agents, (2) imperfect vision-based detection, and (3) potential unknown obstacles in an agent's path.
The notation used is provided in Table~\ref{tab:notation}.

\begin{table}[h]
\renewcommand{\arraystretch}{1.1}
\caption{\centering Notation} \label{tab:notation}
\noindent\resizebox{\columnwidth}{!}{%
\begin{centering}
\begin{tabular}{>{\centering}m{0.15\columnwidth} >{\centering}m{0.85\columnwidth}}
\toprule
\textbf{Symbol} & \textbf{Definition}\tabularnewline
\midrule
$(\cdot)_{k}$ & Variable at time step k. \tabularnewline
\hline
$P$ & Covariance matrix. \tabularnewline
\hline
$H$ & Observation matrix. \tabularnewline
\hline
$F$ & State-transition matrix defined as $e^{At}$, where $A$ is a time-invariant system dynamic matrix.  \tabularnewline
\hline
$\hat{P}_{k|k-1}$ & Estimate of $P$ at time $k$ given observations up to and including time $k-1$. \tabularnewline
\hline
$S$ & Innovation covariance. \tabularnewline
\hline
{$K$} & Optimal Kalman gain. \tabularnewline
\hline
$\boldsymbol{x}$; $\boldsymbol{x}^{o}$ & Ego agent's state; obstacle/peer-agents' state. \tabularnewline
\hline
$R_{\text{max}}$ & Maximum covariance of the observation noise. \tabularnewline
\hline
$\theta$ & Opening angle of the cone that approximates FOV. \tabularnewline
\hline
$f_{FOV}$ & Function that determines if a given point $\boldsymbol{p}$ is in FOV. Defined as $-\cos(\theta/2) + (\boldsymbol{p})_{z}/\lVert \boldsymbol{p} \rVert$. \tabularnewline 
\hline
$R_{FOV}$ & Covariance of observation noise that dynamically changes uncertainty propagation. Defined as $R_{max} \frac{1 - f_{FOV}}{1 + f_{FOV} + \epsilon}$, where $\epsilon$ is a small number. \tabularnewline
\hline
$(\cdot)^{m}$ & Variable in the direction of motion uncertainty formulation. \tabularnewline
\hline
$\boldsymbol{p}_{\text{traj}}$ & Points uniformly sampled from the ego agent’s trajectory. \tabularnewline
\hline
$\overline{\boldsymbol{v}}_{\text{limit}}$ & Upper bound of velocity constraint (constant). \tabularnewline
\hline
$P_{pos}$ & Position covariance matrix. \tabularnewline
\hline
$\boldsymbol{v}_{\text{max}}(P^{m}_{pos})$ & Velocity constraint (dependent of $P^{m}_{pos}$). \tabularnewline
\hline
$\mathcal{F}_w$ & Global world frame.
\tabularnewline \hline
$\mathcal{F}_{\ell_i}$ &
Local and potentially drifted frame of the $i$-th agent---if an agent has perfect localization $\mathcal{F}_{\ell_i} = \mathcal{F}_w$.
\tabularnewline \hline
$\mathcal{F}_i$ &
$i$-th agent's body frame.
\tabularnewline \hline
$\mathcal{F}_c$ &
Camera frame.
\tabularnewline \hline
$X^\beta_\alpha$ &
Pose ($\in SE(d)$ where $d$ is $2$ or $3$) of $\alpha$ in the frame $\mathcal{F}_\beta$ or the rigid body transformation from frame $\mathcal{F}_\alpha$ to frame $\mathcal{F}_\beta$.
\tabularnewline \hline
$p^\alpha_{\text{cent}}$ &
Position of a centroid in the frame $\mathcal{F}_\alpha$
\tabularnewline \hline
$u_{\text{cent}}$, $v_{\text{cent}}$ &
Centroid camera pixel coordinates.
\tabularnewline \hline
{$K_c$} &
Camera intrinsic calibration matrix ($\in \mathbb{R}^{3\times3}$).
\tabularnewline \hline
$\lambda$ &
Positive scalar ($\in \mathbb{R}^+$).
\tabularnewline \hline
$\kappa$ & Number of timesteps to keep a landmark without new measurements before deleting. 
\tabularnewline \hline
$\delta_{i,o}$ &
Time since the $i$-th agent last observed landmark $o$.
\tabularnewline
\bottomrule
\end{tabular}
\par\end{centering}
}
\vspace{-5pt}
\end{table}

\subsection{Uncertainty Propagation for Known Obstacles/Agents}\label{subsec:uncertainty-propagation-for-known-obstacles-and-agents}
Predicting accurate future trajectories of dynamic obstacles from their observation history and accounting for localization errors of other agents is challenging.
Therefore, we need to factor in the uncertainty of these trajectories for robust collision avoidance.

Kamel et al.~\cite{kamel2017robust} employed the Extended Kalman Filter (EKF) for uncertainty propagation of dynamic objects' trajectories:
\begin{equation}\label{eq:uncertainty-propagation-kamel}
    P_{k+1} = F_{k} P_{k} F_{k}^{T}
\end{equation}
The $95\%$ confidence interval error ellipsoid from $P$ is inflated as a minimum allowable acceptable distance between the ego agent and obstacle\textemdash a larger $P$ leads to a bigger boundary.
While effective, this method does not include perception information. We propose a modification to the propagation equation using the prediction and measurement update steps of a linear Kalman Filter:
\begin{equation}\label{eq:uncertainty-propagation}
    \begin{aligned}
        & \hat{P}_{k|k-1} = F_{k} P_{k-1|k-1} F_{k}^{T} \\
        & S_{k} = H_{k} \hat{P}_{k|k-1} H_{k}^{T} + R_{k, FOV}(\boldsymbol{x}_{k}, \boldsymbol{x}^{o}_{k}) \\
        & K_{k} = \hat{P}_{k|k-1} H_{k}^{T} S_{k}^{-1} \\
        & P_{k|k} = (I - K_{k} H_{k}) \hat{P}_{k|k-1} \\
    \end{aligned}
\end{equation}
where the state of each obstacle contains its $\mathbb{R}^{3}$ position, velocity, and acceleration.
If obstacles or other agents are within \ac{fov}, the predicted uncertainty is reduced, simulating the process of obtaining new measurements on the position of the peer agents and obstacles.

\subsection{Uncertainty Propagation to Safely Navigate in Unknown Space}

When navigating in unknown environments, the agent should monitor for unknown obstacles along the direction in which the ego agent is moving (unknown space).
This is obtained by introducing an additional uncertainty propagation, where we modify the innovation covariance formulation in Eq.~\eqref{eq:uncertainty-propagation} as:
\begin{equation}\label{eq:uncertainty-propagation-mv}
    \begin{aligned}
        & S^{m}_{k} = H^{m}_{k} \hat{P}^{m}_{k|k-1} {H^{m}_{k}}^{T} + R^{m}_{k, FOV}(\boldsymbol{x}_{k}, \boldsymbol{p}_{\text{traj}}), \\
    \end{aligned}
\end{equation}
where $R^{m}_{k, FOV}(\boldsymbol{x}_{k}, \boldsymbol{p}_{\text{traj}})$ takes the ego agent's state $\boldsymbol{x}_{k}$ and $\boldsymbol{p}_{\text{traj}}$, a set of points uniformly sampled from the ego agent's future trajectory. 
The propagated uncertainty remains minimal if the points in $\boldsymbol{p}_{\text{traj}}$ are within \ac{fov}. 
As $P^{m}_{pos}$ enlarges, indicating increasing uncertainty along the direction of motion, we impose tighter constraints on the planner's maximum velocity. 
\begin{equation}
        \boldsymbol{v}_{\text{max}, k}(P^{m}_{pos, k}) = \frac{\overline{\boldsymbol{v}}_{\text{limit}}}{\sqrt{\text{diag}(P^{m}_{pos, k})}}
\label{eq:velocity-constraint}
\end{equation}
Note that this operation is done element-wise for the $x$, $y$, and $z$ components.

\subsection{Uncertainty-aware Optimization Formulation}

This section presents the uncertainty-aware optimization formulation, which
leverages the future propagation of $P$ and optimizes the vehicle trajectory over a time horizon, similar to an MPC formulation.
The optimization problem is formulated as follows:
\begin{empheq}[box=\fbox]{flalign*}\label{eq:optimization-cost}
\underset{\makecell{\scriptstyle \mathbf{p}(t), \psi(t), \\ \scriptstyle \boldsymbol{n}_{ij}, d_{ij}, T}}{\boldsymbol{\min}} & \alpha_{\mathbf{j}} \text{\intjerksquared} + \alpha_{\psi} \intddyawsquared + \alpha_T T \\[-2em]
& ~~~~~~ + \alpha_{\boldsymbol{g}_{\mathbf{p}}} \left\Vert \mathbf{p}(T) - \boldsymbol{g}_{\mathbf{p}} \right\Vert ^2 + \alpha_{g_{\psi}} \left\Vert \psi(T) - g_{\psi} \right\Vert ^2  \\
\text{s.t.} ~~& \mathbf{x}(t_{\text{in}})=\mathbf{x}_{\text{in}}, \quad\mathbf{v}(T)=\boldsymbol{0}, \\
& \mathbf{a}(T)=\boldsymbol{0}, \quad\dot{\psi}(T) = 0, \\
& \boldsymbol{n}_{ij}^{T}(P_{pos})\boldsymbol{q}+d_{ij}(P_{pos})<0 \quad \forall \boldsymbol{q} \in \mathcal{Q}_{j} \forall i,j \\
& \text{abs}\left(\boldsymbol{v}\right) \le \boldsymbol{v}_{\text{max}}(P^{m}_{pos}) \quad \forall \boldsymbol{v}\in\mathcal{V}_{j}, \forall j \\
& \text{abs}\left(\boldsymbol{a}_{l}\right) \le \boldsymbol{a}_{\text{max}} \quad \forall l \in L_{\mathbf{p}} \backslash \{n_{\mathbf{p}}-1,n_{\mathbf{p}}\} \\
&\text{abs}\left(\boldsymbol{j}_{l}\right) \le \boldsymbol{j}_{\text{max}} \quad \forall l \in L_{\mathbf{p}}\backslash\{n_{\mathbf{p}}-2,n_{\mathbf{p}}-1,n_{\mathbf{p}}\} \\
& \text{abs}\left(\psi_{l}\right) \le \psi_{\text{max}} \quad  \forall l\in L_{\Psi}\backslash\{n_{\Psi}\} \\
& \text{\Cref{eq:uncertainty-propagation,eq:uncertainty-propagation-mv}} 
\end{empheq}
Our formulation adopted the notation from PANTHER~\cite{tordesillas2021panther}. 
It is essential to note that the variables for the separating plane ($\boldsymbol{n}_{ij}$ and $d_{ij}$) depend on $P_{pos}$, and the velocity constraint is dependent of $P^{m}_{pos}$.
Importantly, this formulation does not necessitate explicit obstacle tracking.
If a trajectory that leads to obstacles being in \ac{fov} of the agent provides a smoother and shorter path, the planner is naturally incentivized to track the obstacles, similarly for the direction of motion.
Our approach is the first to demonstrate implicit obstacle tracking in this way.

\subsection{Multiagent Trajectory Deconfliction}

To share the planned trajectories with other agents, we use our Robust MADER trajectory deconfliction framework~\cite{kondo2023robust} to guarantee safe and collision-free trajectories.
This framework, by introducing (1) a Delay Check step, (2) a two-step trajectory publication scheme, and (3) a trajectory storing-and-checking approach, gives agents the ability to deconflict trajectories asynchronously and without centralized communication.

\section{Frame Alignment Pipeline}\label{sec:frame-alignment-pipeline}

\begin{figure*}[!t]
    \centering
    \includegraphics[width=\textwidth,trim={0 34cm 0 0},clip]{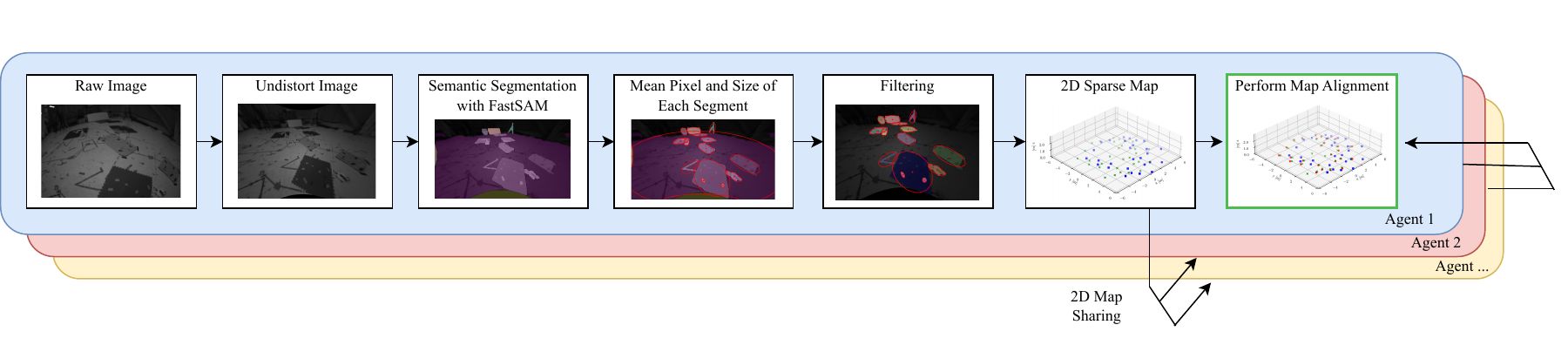}
    \caption{Frame Alignment Pipeline Workflow: The process starts by un-distorting a raw, fisheye image. Next, the pipeline identifies and segments individual objects within the image. We then determine each identified object's centroid, filtering out objects based on size. Once processed, these centroids are projected onto a 2D map, which is used in the frame alignment among agents.}
    \label{fig:frame-alignment-pipeline}
\end{figure*}
To address discrepancies in agents' local frames caused by onboard localization drift, we introduce a pipeline inspired by \cite{peterson2023motlee} where drifted frames of each agent are aligned in real-time and using only distributed communication.
The goal of frame alignment is to find the rigid transformation $X^{\ell_i}_{\ell_j}$ that maps poses from agent $j$'s local frame, $\mathcal{F}_{\ell_j}$, into agent $i$'s local frame, $\mathcal{F}_{\ell_i}$.
To estimate $X^{\ell_i}_{\ell_j}$, each agent $i$ creates maps of static landmarks in the environment and then aligns its map with each of its neighbors $j$.
Maps are created using observations that leverage recent image segmentation techniques~\cite{kirillov2023segany, zhao2023fast}.
This image segmentation allows agents to create maps of previously unencountered objects, removing the need for objects of known classes to exist in the environment.
This workflow is illustrated in Fig.~\ref{fig:frame-alignment-pipeline}.

Map creation begins with detecting segments from an image frame using a generic zero-shot segmentation model \cite{zhao2023fast} and extracting the pixel coordinates $(u_\text{cent}, v_\text{cent})$ segments' centroids. 
Segments are filtered based on size, and then each centroid position is computed with 
\begin{equation}\label{eq:cam-projection}
\begin{aligned}
&p^{\ell_i}_\text{cent} = \boldsymbol{X}^{\ell_i}_{c} (\lambda K_c^{-1} \begin{bmatrix}u_\text{cent} & v_\text{cent} & 1\end{bmatrix}^T). \\
\end{aligned}
\end{equation}
We make the assumption that landmarks are on the ground plane, which constrains the $z$ element of $p^{\ell_i}_\text{cent}$ to be $0$, further defining the value of $\lambda$ and consequently, the $x$ and $y$ components of $p^{\ell_i}_\text{cent}$.
In practice, observed objects will often be 2.5D (i.e., resting on the ground plane but protruding from the plane),
so measurement covariances are stretched along the range direction to represent the uncertainty accurately.

Once object centroids and their associated measurement covariances have been obtained in the drifted world frame, we create maps and perform frame alignment.
Identified centroids are associated with existing map objects using the global nearest neighbor approach \cite{bar1995multitarget}, which formulates a linear assignment problem based on the Mahalanobis distance between new centroid measurements and existing map objects.
New landmarks are created in the map when object measurements cannot be associated with existing landmarks, and landmarks are deleted from the map after $\kappa$ frames of not being seen to eliminate any drift distortion in the map.

Subsequently, pairs of distributed robots align their drifted frames using their landmarks maps.
Before maps can be aligned, pairs of objects within each map must be associated together. 
This is accomplished using CLIPPER~\cite{lusk2021clipper} which performs global data association between objects using geometric consistency.
Once associations have been made, a weight $W(\delta_{i,o}, \delta_{j,o})$ is applied to each pair of associated objects such that $W(\delta_{i,o}, \delta_{j,o}) = ({\delta_{i,o}}{\delta_{j,o}})^{-1}$,
which prioritizes using landmarks with recent observations.
Finally, the transformation $X^{\ell_i}_{\ell_j}$ is computed using a weighted version of Arun's method \cite{arun1987least}, with weights computed using $W(\delta_{i,o}, \delta_{j,o})$.
This transformation is applied to trajectories that an agent receives from its neighbor $j$ to rectify frame misalignment.
\section{Simulation Results}\label{sec:simultation-results}

\subsection{Single-agent Uncertainty-aware Planner Benchmarking}

\begin{figure}[h]
    \centering
    \subcaptionbox{Initial Position $t$=\SI{0}{\s}.~ \label{fig:glance-snapshot1}}{
    \begin{tikzpicture}[every text node part/.style={align=center}]
        \node {\includegraphics[width=0.46\columnwidth, height=0.27\columnwidth, trim={26in 5in 4in 7in}, clip]{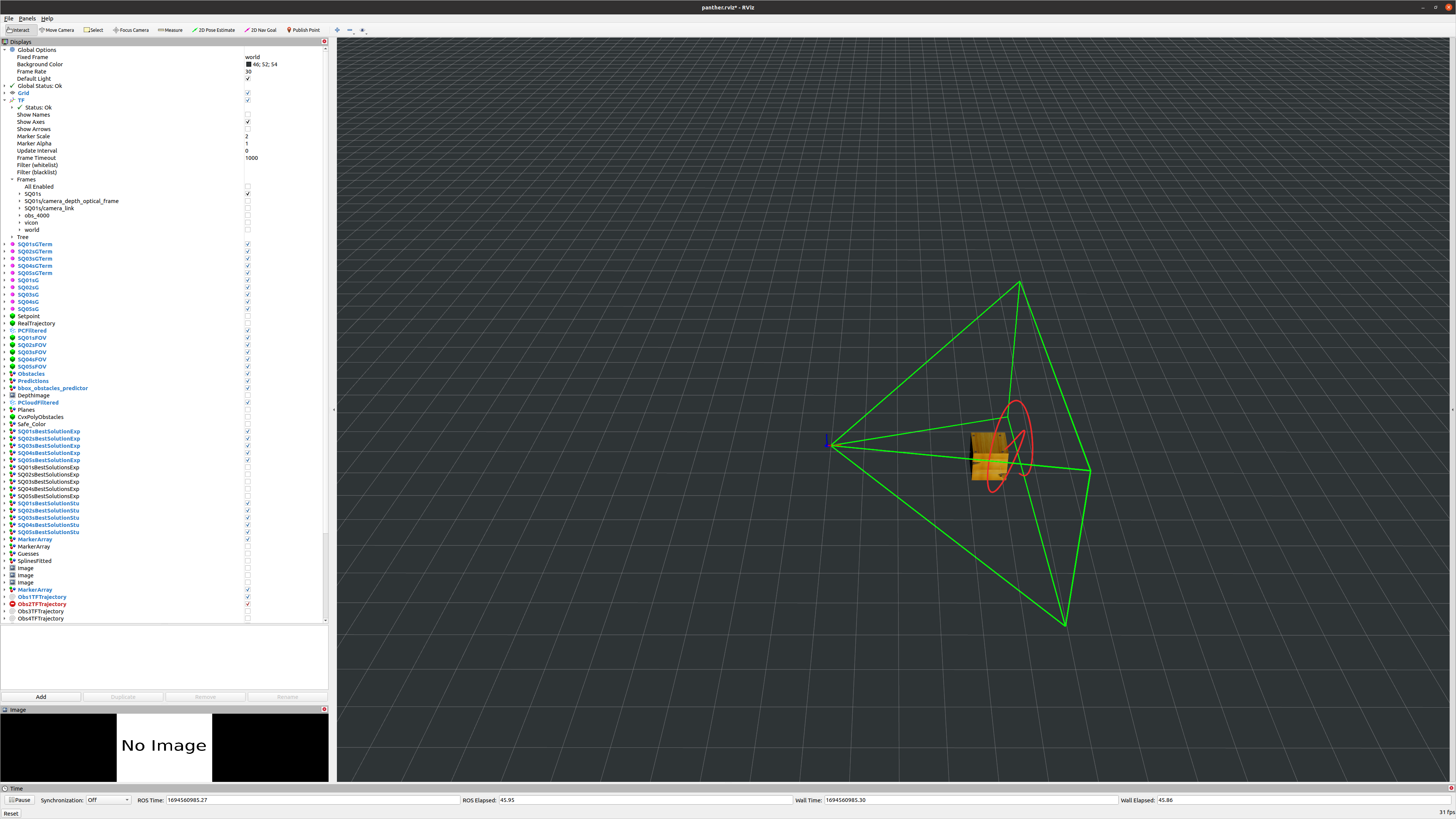}};
        \node [rectangle, fill=white] at (-1.5, 0.8)  {1};
    \end{tikzpicture}}
    \subcaptionbox{Focus on Obst. $t$=\SI{1.0}{\s}.~\label{fig:glance-snapshot2}}{
    \begin{tikzpicture}[every text node part/.style={align=center}]
        \node {\includegraphics[width=0.44\columnwidth, height=0.27\columnwidth, trim={26in 5in 4in 7in}, clip]{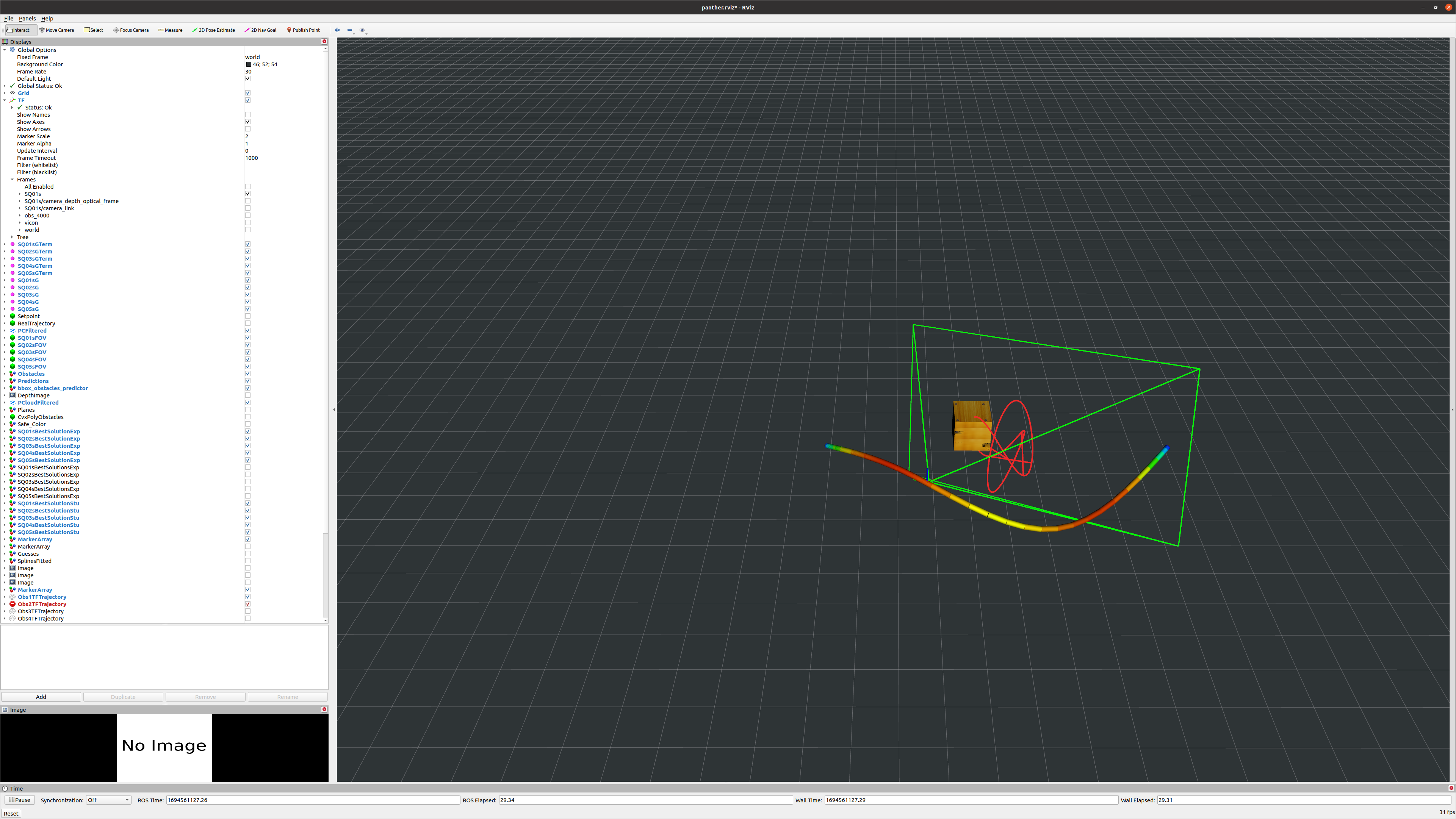}};
        \node [rectangle, fill=white] at (-1.5, 0.8)  {2};
    \end{tikzpicture}}
    \subcaptionbox{Focus on Unknown Space $t$=\SI{2.5}{\s}.~\label{fig:glance-snapshot3}}{
    \begin{tikzpicture}[every text node part/.style={align=center}]
        \node {\includegraphics[width=0.46\columnwidth, height=0.27\columnwidth, trim={26in 5in 4in 7in}, clip]{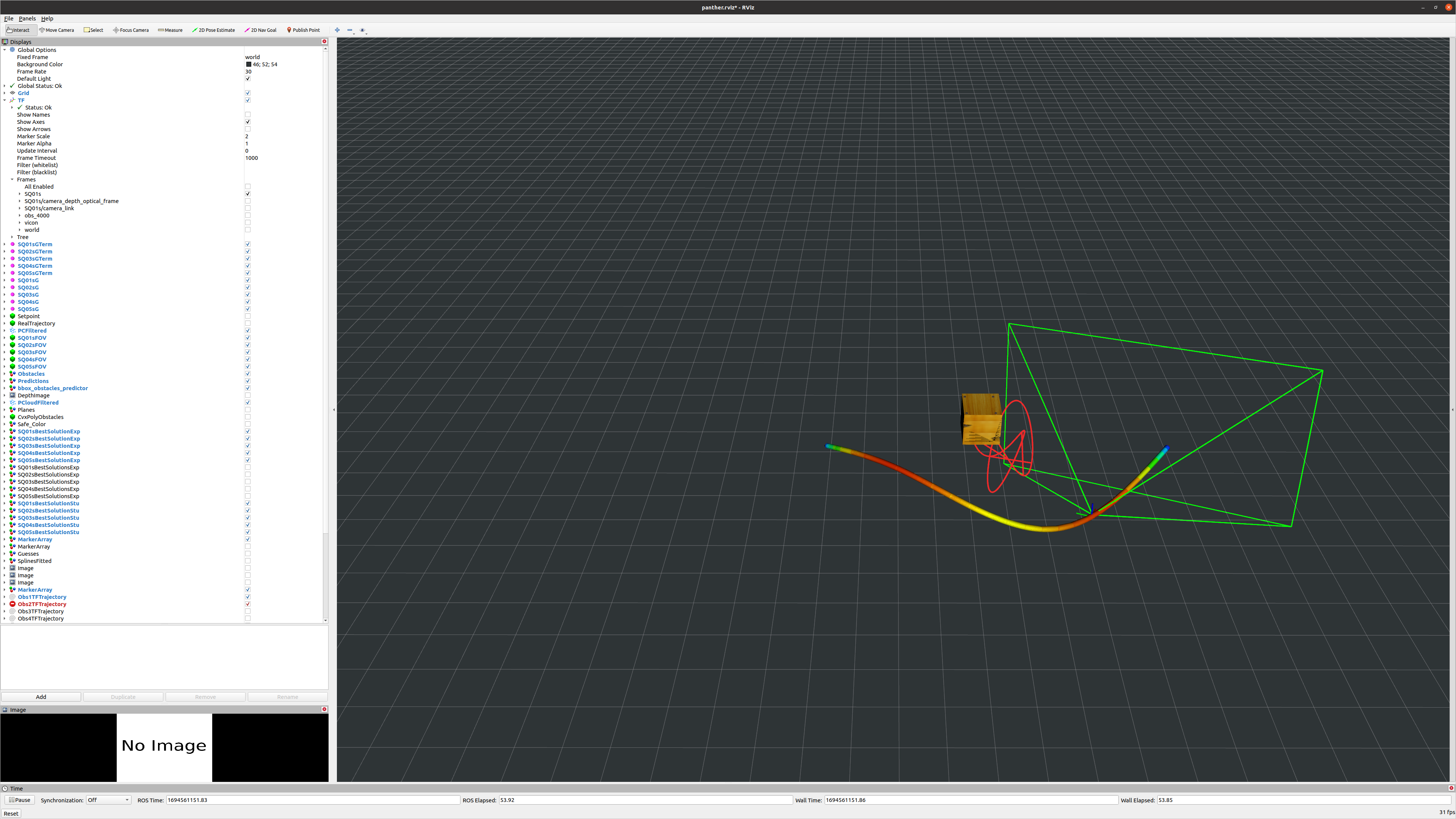}};
        \node [rectangle, fill=white] at (-1.5, 0.8)  {3};
    \end{tikzpicture}}
    \subcaptionbox{End Position $t$=\SI{4.0}{\s}.~\label{fig:glance-snapshot4}}{
    \begin{tikzpicture}[every text node part/.style={align=center}]
    \node {\includegraphics[width=0.44\columnwidth, height=0.27\columnwidth, trim={26in 5in 4in 7in}, clip]{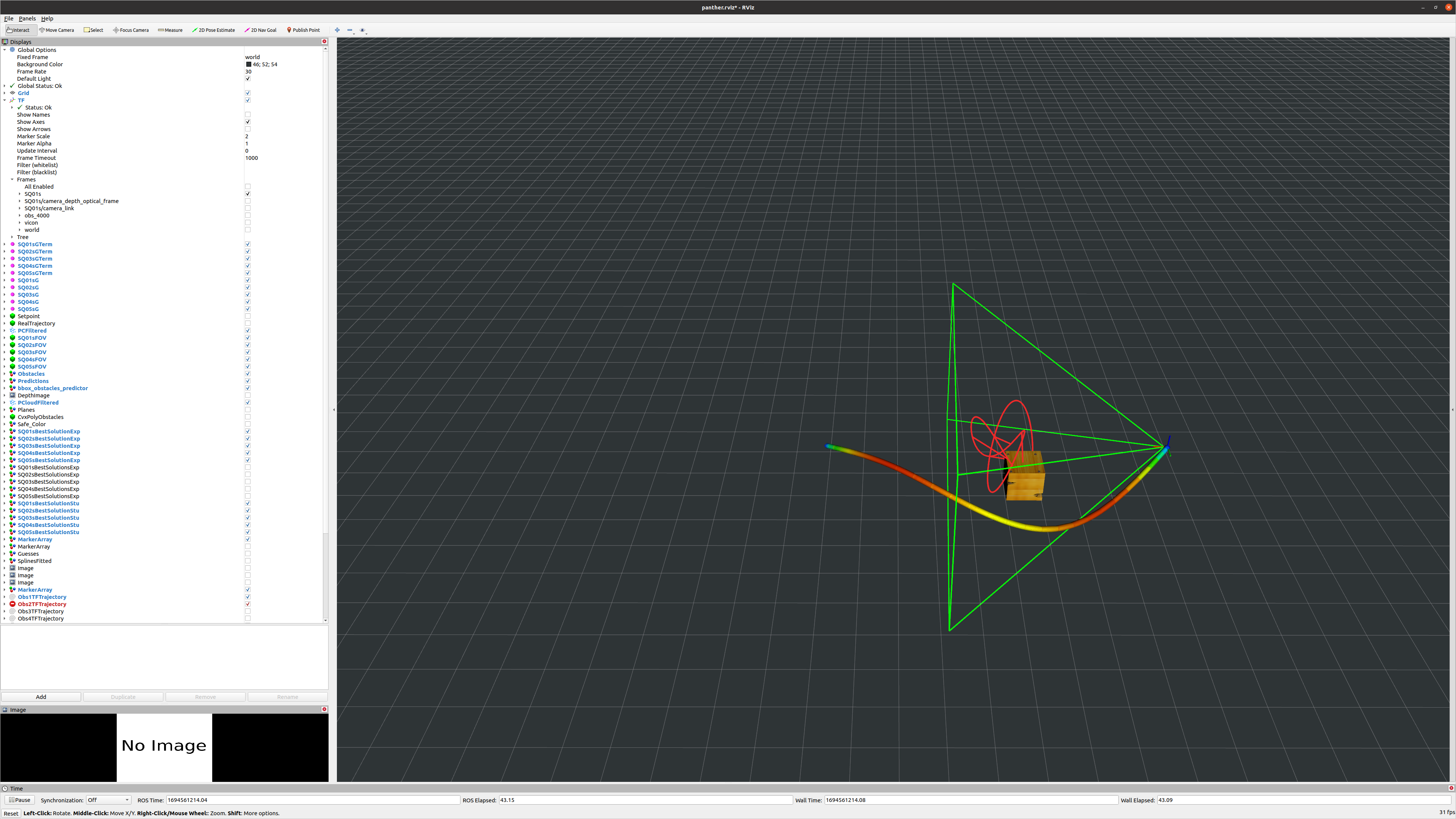}};
    \node [rectangle, fill=white] at (-1.5, 0.8)  {4};
    \end{tikzpicture}}
    \subcaptionbox{Uncertainty Growth: When solving optimization problem, the planner propagates uncertainty of obstacles and unknown space.~\label{fig:uncertainty-growth}}{
    \begin{tikzpicture}[every text node part/.style={align=center}]
    \node {\includegraphics[width=0.9\columnwidth]{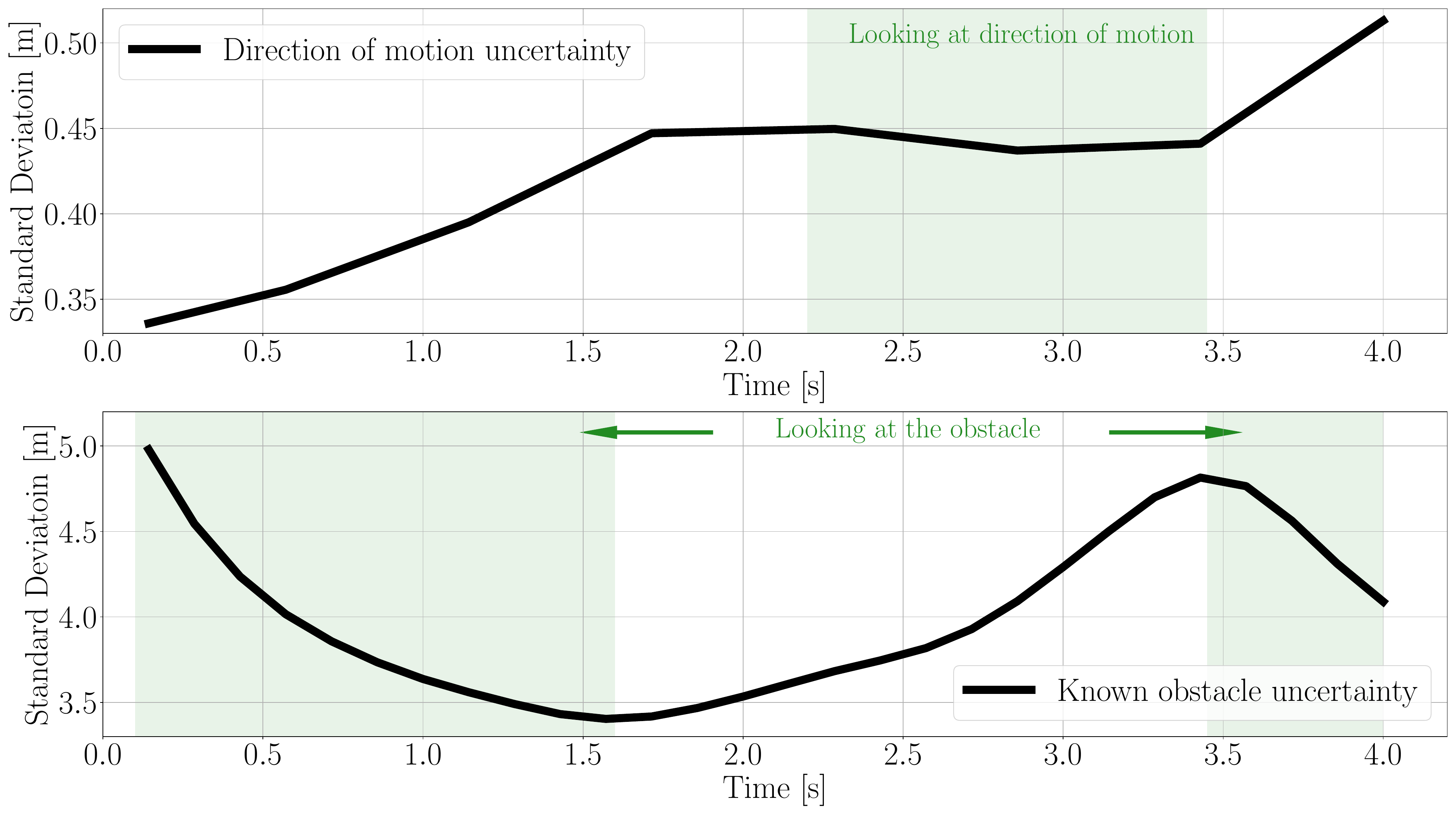}};
    \end{tikzpicture}}
    \caption{PUMA balances reducing uncertainties of (1) known obstacles and (2) potential obstacles in trajectory: The green pyramid represents the \ac*{fov}. 
    The agent navigates around the dynamic obstacle to reach its destination on the opposite side. 
    The trajectory color shows the agent's velocity \textemdash red is fast, and blue slow. \textcolor{red}{}}
    \label{fig:puma-glance}
\end{figure}

This section evaluates the performance of our uncertainty-aware planner through simulations.
We compared our planner with PANTHER*~\cite{tordesillas2023deep}, a state-of-the-art perception-aware planner that has an \textit{explicit obstacle tracking term} in its cost function.
We tested both planners in a single-agent setting, where the agent is tasked to fly through one obstacle that follows a pre-determined trefoil trajectory.
Fig.~\ref*{fig:puma-glance} shows the flight environment, and Table~\ref{tab:sim-ua-planner-single} summarizes the results of 100 flight simulations.
The metrics used to evaluate the performance are as follows:

\begin{enumerate}
    \item Travel time: duration to complete the path.
    \item Computation time: duration to replan at each step.
    \item Number of collisions: the number of collisions that the agent experiences.
    \item Known obstacle \ac{fov} rate: the percentage of time that the agent keeps known obstacles within its \ac{fov}.
    \item Known obstacle continuous \ac{fov} detection: Consecutive time that an obstacle is continuously kept within the \ac{fov} of the agent.
    \item Unknown space \ac{fov} rate: the percentage of time that the agent keeps unknown space (direction of motion) within its \ac{fov}.
    \item Unknown space continuous \ac{fov} detection: consecutive time that unknown space (direction of motion) is continuously kept within the \ac{fov} of the agent.
\end{enumerate}

\begin{table}[h]
    \renewcommand{\arraystretch}{1}
    \scriptsize
    \begin{centering}
    \caption{\centering Single Agent Uncertainty-aware Planner Benchmarking \textemdash \ 100 simulations \label{tab:sim-ua-planner-single}}
    \resizebox{1.0\columnwidth}{!}{
    \begin{tabular}{>{\centering\arraybackslash}m{0.17\columnwidth} >{\centering\arraybackslash}m{0.1\columnwidth} >{\centering\arraybackslash}m{0.1\columnwidth} >{\centering\arraybackslash}m{0.1\columnwidth} >{\centering\arraybackslash}m{0.12\columnwidth} >{\centering\arraybackslash}m{0.12\columnwidth} >{\centering\arraybackslash}m{0.12\columnwidth} >{\centering\arraybackslash}m{0.12\columnwidth} }
    \toprule 
    \multirow{2}[5]{*}{\textbf{Planner}} & \multirow{2}[5]{*}{\makecell{\textbf{Travel} \\ \textbf{Time [s]}}} & \multirow{2}[5]{*}{\makecell{\textbf{Comp.} \\ \textbf{Time [ms]}}} & \multirow{2}[5]{*}{\textbf{\# Colls.}} & \multicolumn{2}{c}{\textbf{Known Obst.}} & \multicolumn{2}{c}{\textbf{Unknown Space}} \tabularnewline
    &&&& \textbf{\ac{fov} Rate [\%]} & \textbf{Conti. \ac{fov} Dets. [s]} & \textbf{\ac{fov} Rate [\%]} & \textbf{Conti. \ac{fov} Dets. [s]} \tabularnewline
    \midrule
    \textbf{PUMA} (proposed) & 5.2 & 5712 & \textbf{\textcolor{ForestGreen}{0}} & \textbf{\textcolor{orange}{62.8}} & 2.4 & \textbf{\textcolor{ForestGreen}{13.0}} & \textbf{\textcolor{ForestGreen}{0.59}} \tabularnewline
    \midrule
    \textbf{PANTHER*}~\cite{tordesillas2023deep} & 4.5 & 1891 & \textbf{\textcolor{ForestGreen}{0}} & \textbf{\textcolor{ForestGreen}{100}} & 4.5 & \textbf{\textcolor{red}{0.1}} & \textbf{\textcolor{red}{0.0}} \tabularnewline
    \bottomrule
    \end{tabular}}
    \par\end{centering}
\end{table}

Table~\ref{tab:sim-ua-planner-single} highlights a key distinction between PUMA and PANTHER*~\cite{tordesillas2023deep}. 
Specifically, PUMA places emphasis on perceiving unknown spaces while concurrently maintaining awareness of known obstacles. 
PUMA retains perception of the known obstacles for \SI{62.8}{\%} of the flight time while focusing on the unknown space (direction of motion) for \SI{13.0}{\%}.
In contrast, PANTHER*, due to the \textit{explicit obstacle tracking term} in its cost function, remains focused on known obstacles for \SI{100}{\%} of the flight duration, giving no attention to unknown spaces, shown by a rate of \textcolor{red}{\SI{0.1}{\%}}.

Fig.~\ref*{fig:puma-glance} illustrates PUMA's capability to balance uncertainties about obstacles and unknown spaces. 
Specifically, at $t$=\SI{1.0}{\s}, as depicted in Fig.\ref{fig:glance-snapshot2}, the agent focuses on the known obstacle, decreasing its uncertainty while uncertainty about unknown space goes up.
Conversely, at $t$=\SI{2.5}{\s}, the agent's attention shifts to unknown spaces it approaches, reducing its uncertainty as indicated in Fig.~\ref{fig:uncertainty-growth}.
This dynamic management of uncertainties allows the agent to navigate unknown environments safely.

While PUMA takes more computation time than PANTHER*, the imitation learning approach used for Deep-PANTHER~\cite{tordesillas2023deep}, can be used in future work to significantly reduce the computation time of PUMA while maintaining very similar levels of performance.

\subsection{Frame Alignment Pipeline Evaluation}\label{subsec:frame-alignment-pipeline-evaluation}

This section evaluates our image segmentation-based frame-alignment pipeline. 
We use simulated environments involving two flying agents and artificially introduce either constant bias or linear drift in translation and yaw to the state estimation of one agent.
Our pipeline effectively detects these misalignments and estimates the relative states.

\subsubsection{Simulation Environments}

\begin{figure}[h]
    \centering
    \subcaptionbox{Pads Environment.\label{fig:frame-align-pads-env}}{
    \begin{tikzpicture}[every text node part/.style={align=center}]
    \node {\includegraphics[width=0.45\columnwidth]{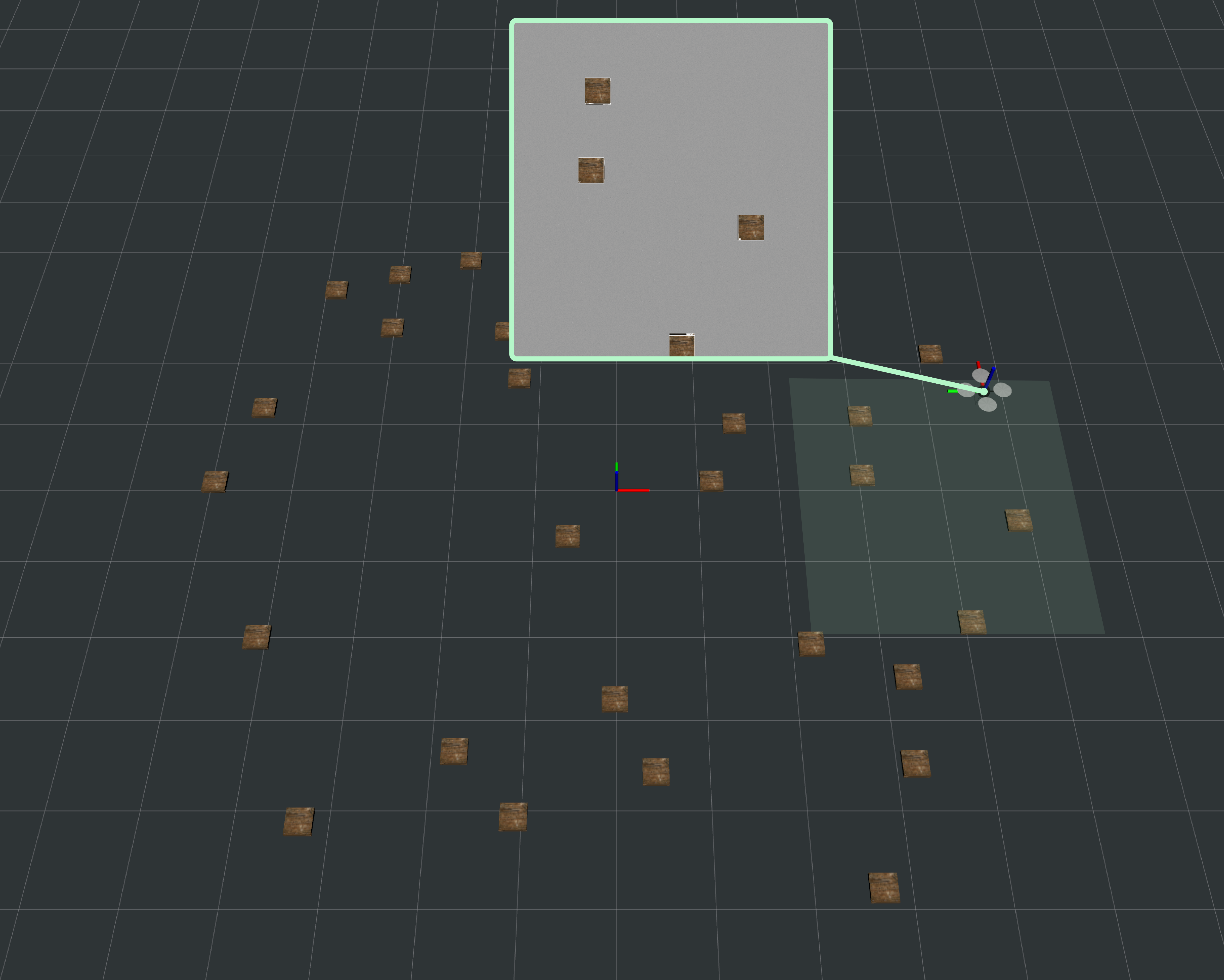}};
    \end{tikzpicture}}
    \subcaptionbox{Random Objects Environment.\label{fig:frame-align-random-objects-env}}{
    \begin{tikzpicture}[every text node part/.style={align=center}]
    \node {\includegraphics[width=0.45\columnwidth]{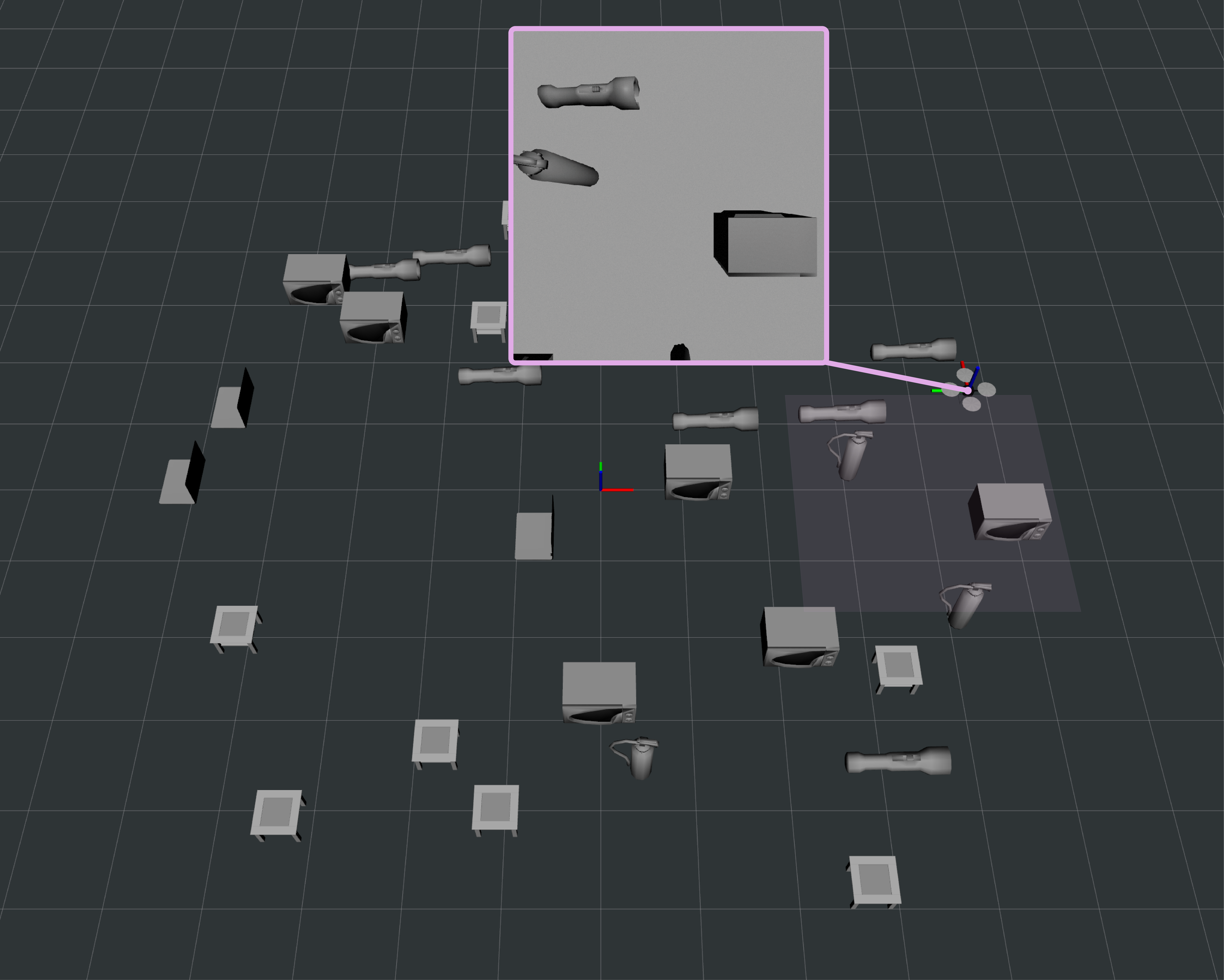}};
    \end{tikzpicture}}
    \caption{Pipeline Evaluation Simulation Environments.}
    \label{fig:simulation-environments}
    \vspace{-1em}
\end{figure}

We tested the pipeline using two sets of objects: flat pads and random objects.
As mentioned in Section~\ref{sec:frame-alignment-pipeline}, our pipeline assumes that objects are on the ground.
Consequently, it performs better with flat pads depicted in Fig.~\ref{fig:frame-align-pads-env}. 
However, to further evaluate its robustness, we also examined it with various randomly chosen objects, as illustrated in Fig.\ref{fig:frame-align-random-objects-env}.

For the simulations, we employed the camera model from Intel\textsuperscript{\textregistered} RealSense\textsuperscript{TM} T265. 
The raw images, as displayed in Figs.~\ref{fig:frame-align-pads-env} and ~\ref{fig:frame-align-random-objects-env}, were fed directly into our segmentation-based pipeline.

\subsubsection{Simulation Results}

We evaluate our pipeline by considering two different trajectories: (1) agents following the same circular trajectory and (2) \ac{poc} trajectory.
For each trajectory, we tested three different drift scenarios: (1) no drift, (2) constant frame offset, and (3) linearly added drift. We introduced these drifts and offsets to vehicle 1 in a) translation only, b) yaw only, and c) both translation and yaw. 
When applied, the constant bias consists of \SI{1}{m} along the $x$ and $y$ axes and \SI{10}{deg} in yaw, and linear drift accumulates at a rate of \SI{.05}{m/s} along $x$ and $y$ and \SI{.05}{deg/s} in yaw.
Accurately aligning frames is more difficult when drift accumulates linearly than when there is only a constant bias, and drift in both translation and yaw presents a more difficult scenario than drift in only translation or only yaw.

\begin{figure}[t]
\centering
    \includegraphics[width=\columnwidth, trim=3.8in 2in 4.5in 1in, clip]{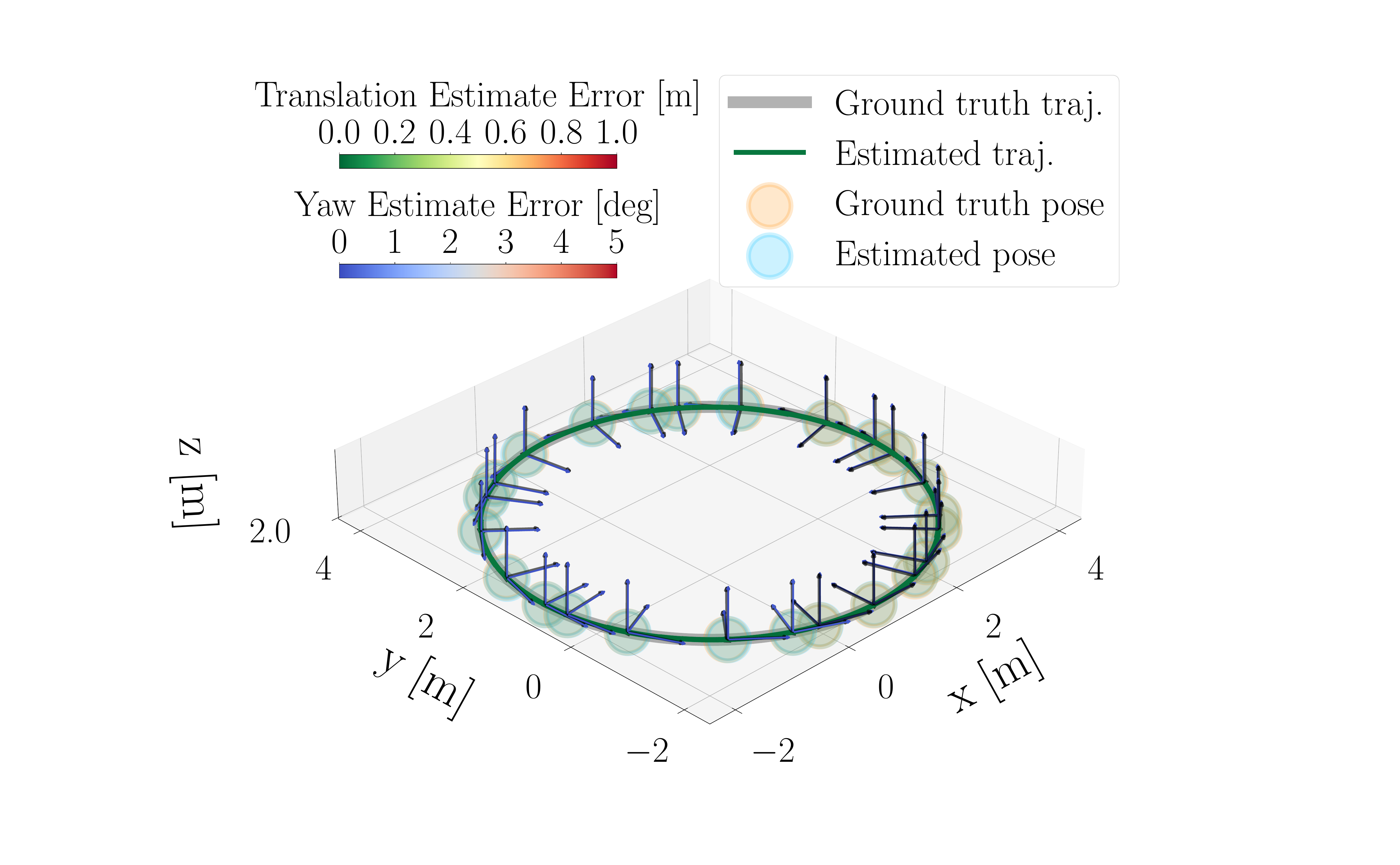}
    \caption{Frame alignment quality in Case 3 (difficulty: \Easy): Trajectory color shows the estimation error associated with it and the color of the coordinate frame attached to vehicle 1 shows the yaw estimation error. The pipeline successfully estimates the drifted state of vehicle~1 with minimal errors (Table~\ref{tab:sim-benchmarking-pads}). The coordinate frames are captured at intervals of \SI{5}{\s}.}
    \label{fig:pcc-map-tracking-quality}
\end{figure}

We first showcase the frame alignment results of our pipeline under artificially introduced constant drifts in Fig.~\ref{fig:pcc-map-tracking-quality}.
The pipeline first identifies the pads in the environment and places them into a 2D map, and it performs frame alignment in real-time and estimates vehicle 1's drifted state.
The estimated state and ground truth values in Fig.~\ref{fig:pcc-map-tracking-quality} show that the pipeline successfully estimates the drifted state of vehicle 1 with minimal errors. Tables~\ref{tab:sim-benchmarking-pads} and Table~\ref{tab:sim-benchmarking-daily-commodity} summarize the results of our simulations.
The tables are organized as follows: (1) Columns 1--4 show the trajectory and drift types, (2) Columns 5--7 show the drift values, (3) the last 6 columns show the mean and standard deviation of the $x$, $y$, and yaw errors, respectively. Table~\ref{tab:sim-benchmarking-pads}~\ref{tab:multiagent-sim-benchmarking-random} shows that our pipeline successfully estimates the drifted state of vehicle 1 in all cases.
It is apparent that the error increases for the more difficult cases, but even in the most challenging case, the mean yaw error is less than 2.5$^\circ$, and the mean position error is less than \SI{0.21}{\m}.
Note that the random objects are 3D objects and, thus, are not flat on the ground. 
Therefore, it is challenging to detect them; however, our pipeline still successfully estimates the drifted state of vehicle 1.

\begin{table}[t]
    \renewcommand{\arraystretch}{1}
    \scriptsize
    \begin{centering}
    \caption{\centering Simulation Benchmarking 1 \textemdash \ Object Type: Pads}
    \label{tab:sim-benchmarking-pads}
    \resizebox{1.0\columnwidth}{!}{
    \begin{tabular}{>{\centering\arraybackslash}m{0.03\columnwidth} >{\centering\arraybackslash}m{0.15\columnwidth} >{\centering\arraybackslash}m{0.05\columnwidth} >{\centering\arraybackslash}m{0.15\columnwidth} >{\centering\arraybackslash}m{0.12\columnwidth} >{\centering\arraybackslash}m{0.08\columnwidth} >{\centering\arraybackslash}m{0.07\columnwidth} >{\centering\arraybackslash}m{0.07\columnwidth} >{\centering\arraybackslash}m{0.07\columnwidth} >{\centering\arraybackslash}m{0.07\columnwidth} >{\centering\arraybackslash}m{0.07\columnwidth} }
        \toprule
        \multicolumn{5}{c}{\textbf{Environment Settings}} & \multicolumn{6}{c}{\textbf{Results}} \tabularnewline
        \cmidrule(lr){1-5}
        \cmidrule(lr){6-11}
        \multirow{2}[2]{*}{\textbf{Case}} & \multirow{2}[2]{*}{\textbf{Difficulty}} & \multirow{2}[2]{*}{\textbf{Traj.}} & \multirow{2}[2]{*}{\shortstack{\textbf{Added} \\ \textbf{Error Type}}} & \multirow{2}[2]{*}{\shortstack{\textbf{Translation/} \\ \textbf{Yaw}}} & \multicolumn{2}{c}{\textbf{X Err. [m]}} & \multicolumn{2}{c}{\textbf{Y Err. [m]}} & \multicolumn{2}{c}{\textbf{Yaw Err. [deg]}} \tabularnewline
        &&&&& \textbf{Mean} & \textbf{Std.} & \textbf{Mean} & \textbf{Std.} & \textbf{Mean} & \textbf{Std.} \tabularnewline
        \cmidrule(lr){1-5}
        \cmidrule(lr){6-11}
        1 & \VeryEasy & Circle & None & None & 0.0 & 0.02 & 0.0 & 0.02 & 0.01 & 0.1 \tabularnewline
    \cmidrule(lr){1-5}
    \cmidrule(lr){6-11}
    2 & \Easy & \ac{poc} & None & None & 0.09 & 0.05 & 0.07 & 0.07 & 0.14 & 0.23 \tabularnewline
    \cmidrule(lr){1-5}
    \cmidrule(lr){6-11}
    3 & \Easy & \multirow{4}[0]{*}{Circle} & Const. & Translation & 0.0 & 0.02 & 0.0 & 0.02 & 0.02 & 0.09 \tabularnewline 
    4 & \Easy &                            & Const. & Yaw & 0.0 & 0.01 & 0.0 & 0.02 & 0.05 & 0.1 \tabularnewline
    5 & \Moderate &                        & Const. & Both & 0.0 & 0.02 & 0.0 & 0.02 & 0.05 & 0.13 \tabularnewline
    \cmidrule(lr){1-5}
    \cmidrule(lr){6-11}
    6 & \Easy & \multirow{4}[0]{*}{\ac{poc}} & Const. & Translation & 0.13 & 0.05 & 0.03 & 0.03 & 0.14 & 0.28 \tabularnewline
    7 & \Easy &                          & Const. & Yaw & 0.09 & 0.07 & 0.06 & 0.1 & 0.75 & 0.97 \tabularnewline
    8 & \Moderate &                      & Const. & Both & 0.09 & 0.07 & 0.02 & 0.08 & 1.14 & 1.52 \tabularnewline
    \cmidrule(lr){1-5}
    \cmidrule(lr){6-11}
    9 & \Moderate & \multirow{4}[0]{*}{Circle} & Linear & Translation & 0.01 & 0.11 & 0.02 & 0.11 & 2.42 & 0.14 \tabularnewline
    10 & \Moderate &                            & Linear  & Yaw & 0.0 & 0.01 & 0.0 & 0.01 & 0.52 & 0.12 \tabularnewline
    11 & \Hard     &                            & Linear  & Both & 0.01 & 0.1 & 0.03 & 0.1 & 1.92 & 0.14 \tabularnewline
    \cmidrule(lr){1-5}
    \cmidrule(lr){6-11}
    12 & \Hard & \multirow{4}[0]{*}{\ac{poc}} & Linear & Translation & 0.13 & 0.11 & 0.05 & 0.1 & 2.7 & 0.22 \tabularnewline
    13 & \Hard &                        & Linear & Yaw & 0.11 & 0.02 & 0.02 & 0.01 & 0.54 & 0.23 \tabularnewline
    14 & \VeryHard &                        & Linear & Both & 0.14 & 0.09 & 0.04 & 0.09 & 2.22 & 0.19 \tabularnewline
    \bottomrule
\end{tabular}}
\vspace{-10pt}
\par\end{centering}
\end{table}

\subsection{Multiagent Uncertainty-Aware Planner Evaluation on Frame Alignment Pipeline}

\begin{figure}[t]
    \centering
    \includegraphics[width=\columnwidth, trim=3.8in 2in 4.5in 1.5in, clip]{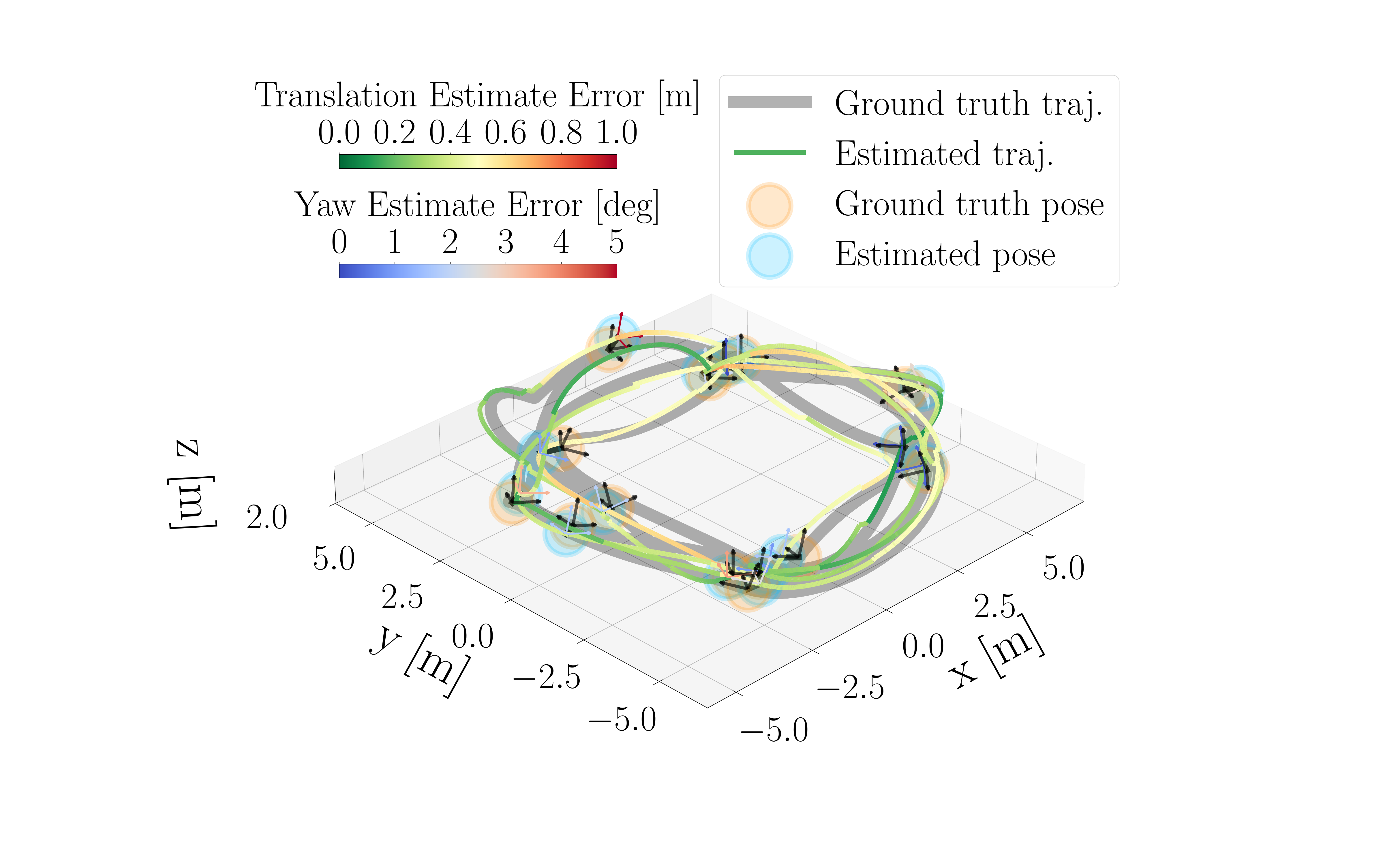}
    \caption{Tracking Quality in Case 42 (difficulty: \VeryHard): Even in the hardest case [linearly increasing drift in x, y, and yaw, random objects, PUMA trajectory], our pipeline successfully estimates the drifted state of vehicle~1. The coordinate frames are captured at intervals of \SI{5}{\s}.}
    \label{fig:multiagent-rlv-tracking-quality}
    \vspace{-1em}
\end{figure}

This section tests the uncertainty-aware planner presented in \Cref{sec:ua-trajectory}, integrated with the frame alignment pipeline discussed in \Cref{sec:frame-alignment-pipeline}.
Following the same pattern as in \Cref{sec:frame-alignment-pipeline}, both constant and linearly increasing drifts were introduced in environments with pads and random objects.
A distinguishing feature here, as opposed to \Cref{subsec:frame-alignment-pipeline-evaluation}, is that the agents' trajectories are not simple circles or \ac{poc} trajectories. 
Instead, they are shaped by the uncertainty-aware planner, introducing an extra layer of complexity to the frame alignment task.
Yet, despite this additional complexity in the vehicle motion, our pipeline successfully estimates frame misalignment.
As \Cref{tab:multiagent-sim-benchmarking-pads,tab:multiagent-sim-benchmarking-random} indicate, the pipeline reliably approximates the drifted state of vehicle 1 in every scenario. 
\Cref{fig:multiagent-pcc-tracking-quality,fig:multiagent-rlv-tracking-quality} further depicts how our image segmentation-based method estimate drifts in a multiagent environment.

\begin{table}[t]
    \renewcommand{\arraystretch}{1}
    \scriptsize
    \begin{centering}
    \caption{\centering Simulation Benchmarking 2 \textemdash \ Object Type: Random}
    \label{tab:sim-benchmarking-daily-commodity}
    \resizebox{1.0\columnwidth}{!}{
        \begin{tabular}{>{\centering\arraybackslash}m{0.03\columnwidth} >{\centering\arraybackslash}m{0.15\columnwidth} >{\centering\arraybackslash}m{0.05\columnwidth} >{\centering\arraybackslash}m{0.15\columnwidth} >{\centering\arraybackslash}m{0.12\columnwidth} >{\centering\arraybackslash}m{0.08\columnwidth} >{\centering\arraybackslash}m{0.07\columnwidth} >{\centering\arraybackslash}m{0.07\columnwidth} >{\centering\arraybackslash}m{0.07\columnwidth} >{\centering\arraybackslash}m{0.07\columnwidth} >{\centering\arraybackslash}m{0.07\columnwidth} }
            \toprule
            \multicolumn{5}{c}{\textbf{Environment Settings}} & \multicolumn{6}{c}{\textbf{Results}} \tabularnewline
            \cmidrule(lr){1-5}
            \cmidrule(lr){6-11}
            \multirow{2}[2]{*}{\textbf{Case}} & \multirow{2}[2]{*}{\textbf{Difficulty}} & \multirow{2}[2]{*}{\textbf{Traj.}} & \multirow{2}[2]{*}{\shortstack{\textbf{Added Offset/} \\ \textbf{Drift Type}}} & \multirow{2}[2]{*}{\shortstack{\textbf{Translation/} \\ \textbf{Yaw}}} & \multicolumn{2}{c}{\textbf{X Err. [m]}} & \multicolumn{2}{c}{\textbf{Y Err. [m]}} & \multicolumn{2}{c}{\textbf{Yaw Err. [deg]}} \tabularnewline
            &&&&& \textbf{Mean} & \textbf{Std.} & \textbf{Mean} & \textbf{Std.} & \textbf{Mean} & \textbf{Std.} \tabularnewline
            \cmidrule(lr){1-5}
            \cmidrule(lr){6-11}
    15 & \Easy & Circle & None & None & 0.01 & 0.02 & 0.01 & 0.02 & 0.16 & 0.3 \tabularnewline
    \cmidrule(lr){1-5}
    \cmidrule(lr){6-11}
    16 & \Moderate & \ac{poc} & None & None & 0.05 & 0.0 & 0.03 & 0.0 & 0.76 & 0.0 \tabularnewline
    \cmidrule(lr){1-5}
    \cmidrule(lr){6-11}
    17 & \Moderate & \multirow{4}[0]{*}{Circle} & Const. & Translation & 0.0 & 0.02 & 0.0 & 0.02 & 0.09 & 0.26 \tabularnewline
    18 & \Moderate &                            & Const. & Yaw & 0.0 & 0.02 & 0.01 & 0.02 & 0.1 & 0.26 \tabularnewline
    19 & \Hard &                                & Const. & Both & 0.0 & 0.03 & 0.01 & 0.02 & 0.05 & 0.34 \tabularnewline
    \cmidrule(lr){1-5}
    \cmidrule(lr){6-11}
    20 & \Moderate & \multirow{4}[0]{*}{\ac{poc}} & Const. & Translation & 0.1 & 0.08 & 0.04 & 0.08 & 0.34 & 0.77 \tabularnewline
    21 & \Moderate &                                    & Const. & Yaw & 0.16 & 0.11 & 0.07 & 0.09 & 1.1 & 2.3 \tabularnewline
    22 & \Hard &                                    & Const. & Both & 0.14 & 0.06 & 0.04 & 0.04 & 0.84 & 0.69 \tabularnewline
    \cmidrule(lr){1-5}
    \cmidrule(lr){6-11}
    23 & \Hard & \multirow{4}[0]{*}{Circle} & Linear & Translation & 0.01 & 0.11 & 0.03 & 0.12 & 2.49 & 0.28 \tabularnewline
    24 & \Hard &                            & Linear & Yaw & 0.01 & 0.02 & 0.0 & 0.03 & 0.63 & 0.24 \tabularnewline
    25 & \VeryHard &                        & Linear & Both & 0.0 & 0.1 & 0.02 & 0.1 & 2.02 & 0.25 \tabularnewline
    \cmidrule(lr){1-5}
    \cmidrule(lr){6-11}
    26 & \VeryHard & \multirow{4}[0]{*}{\ac{poc}} & Linear & Translation & 0.11 & 0.24 & 0.01 & 0.13 & 1.4 & 1.21 \tabularnewline
    27 & \VeryHard &                                & Linear & Yaw & 0.05 & 0.3 & 0.08 & 0.21 & 1.51 & 1.05 \tabularnewline
    28 & \VeryHard &                                & Linear & Both & 0.12 & 0.12 & 0.02 & 0.16 & 1.68 & 1.16 \tabularnewline
    \bottomrule
    \end{tabular}}
    \par\end{centering}
    \renewcommand{\arraystretch}{1}
    \scriptsize
    \begin{centering}
    \caption{\centering Multiagent Simulation Benchmarking \textemdash \ PUMA on Frame Alignment Pipeline on Pads}
    \label{tab:multiagent-sim-benchmarking-pads}
    \resizebox{1.0\columnwidth}{!}{
    \begin{tabular}{>{\centering\arraybackslash}m{0.03\columnwidth} >{\centering\arraybackslash}m{0.15\columnwidth} >{\centering\arraybackslash}m{0.15\columnwidth} 
        >{\centering\arraybackslash}m{0.12\columnwidth} >{\centering\arraybackslash}m{0.08\columnwidth} >{\centering\arraybackslash}m{0.07\columnwidth} >{\centering\arraybackslash}m{0.07\columnwidth} 
        >{\centering\arraybackslash}m{0.07\columnwidth} >{\centering\arraybackslash}m{0.07\columnwidth} >{\centering\arraybackslash}m{0.07\columnwidth} >{\centering\arraybackslash}m{0.08\columnwidth} }
    \toprule
    \multicolumn{4}{c}{\textbf{Environment Settings}} & \multicolumn{7}{c}{\textbf{Results}} \tabularnewline
    \cmidrule(lr){1-4}
    \cmidrule(lr){5-11}
    \multirow{2}[2]{*}{\textbf{Case}} & \multirow{2}[2]{*}{\textbf{Difficulty}} & \multirow{2}[2]{*}{\shortstack{\textbf{Added Offset/} \\ \textbf{Drift Type}}} & 
    \multirow{2}[2]{*}{\shortstack{\textbf{Translation/} \\ \textbf{Yaw}}} & \multicolumn{2}{c}{\textbf{X Err. [m]}} & \multicolumn{2}{c}{\textbf{Y Err. [m]}} & \multicolumn{2}{c}{\textbf{Yaw Err. [deg]}} & \multirow{2}[2]{*}{\textbf{\# Colls.}} \tabularnewline
    &&&& \textbf{Mean} & \textbf{Std.} & \textbf{Mean} & \textbf{Std.} & \textbf{Mean} & \textbf{Std.} & \tabularnewline
    \cmidrule(lr){1-4}\cmidrule(lr){5-11}
    29 & \Moderate & None & None & 0.0 & 0.04 & 0.0 & 0.03 & 0.15 & 0.45 & 0 \tabularnewline
    \cmidrule(lr){1-4}\cmidrule(lr){5-11}
    30 & \Moderate & Const. & Translation & 0.02 & 0.06 & 0.01 & 0.04 & 0.04 & 0.47 & 0 \tabularnewline
    31 & \Moderate & Const. & Yaw & 0.02 & 0.04 & 0.01 & 0.03 & 0.25 & 0.34 & 0 \tabularnewline
    32 & \Hard     & Const. & Both & 0.02 & 0.03 & 0.0 & 0.03 & 0.11 & 0.28 & 0 \tabularnewline
    \cmidrule(lr){1-4}\cmidrule(lr){5-11}
    33 & \VeryHard & Linear & Translation & 0.02 & 0.2 & 0.04 & 0.23 & 0.01 & 1.78 & 0 \tabularnewline
    34 & \VeryHard & Linear & Yaw & 0.01 & 0.07 & 0.02 & 0.06 & 0.7 & 0.38 & 0 \tabularnewline
    35 & \VeryHard & Linear & Both & 0.18 & 0.19 & 0.05 & 0.17 & 1.31 & 0.88 & 0 \tabularnewline
    \bottomrule
    \end{tabular}}
    \par\end{centering}
    \renewcommand{\arraystretch}{1}
    \scriptsize
    \begin{centering}
    \caption{\centering Multiagent Simulation Benchmarking \textemdash \ PUMA on Frame Alignment Pipeline on Random Objects}
    \label{tab:multiagent-sim-benchmarking-random}
    \resizebox{1.0\columnwidth}{!}{
        \begin{tabular}{>{\centering\arraybackslash}m{0.03\columnwidth} >{\centering\arraybackslash}m{0.15\columnwidth} >{\centering\arraybackslash}m{0.15\columnwidth} 
            >{\centering\arraybackslash}m{0.12\columnwidth} >{\centering\arraybackslash}m{0.08\columnwidth} >{\centering\arraybackslash}m{0.07\columnwidth} >{\centering\arraybackslash}m{0.07\columnwidth} 
            >{\centering\arraybackslash}m{0.07\columnwidth} >{\centering\arraybackslash}m{0.07\columnwidth} >{\centering\arraybackslash}m{0.07\columnwidth} >{\centering\arraybackslash}m{0.08\columnwidth} }
        \toprule
        \multicolumn{4}{c}{\textbf{Environment Settings}} & \multicolumn{7}{c}{\textbf{Results}} \tabularnewline
        \cmidrule(lr){1-4}\cmidrule(lr){5-11}
        \multirow{2}[2]{*}{\textbf{Case}} & \multirow{2}[2]{*}{\textbf{Difficulty}} & \multirow{2}[2]{*}{\shortstack{\textbf{Added Offset/} \\ \textbf{Drift Type}}} & 
        \multirow{2}[2]{*}{\shortstack{\textbf{Translation/} \\ \textbf{Yaw}}} & \multicolumn{2}{c}{\textbf{X Err. [m]}} & \multicolumn{2}{c}{\textbf{Y Err. [m]}} & \multicolumn{2}{c}{\textbf{Yaw Err. [deg]}} & \multirow{2}[2]{*}{\textbf{\# Colls.}} \tabularnewline
        &&&& \textbf{Mean} & \textbf{Std.} & \textbf{Mean} & \textbf{Std.} & \textbf{Mean} & \textbf{Std.} & \tabularnewline
        \cmidrule(lr){1-4}\cmidrule(lr){5-11}
    36 & \Hard & None & None & 0.0 & 0.05 & 0.02 & 0.06 & 0.38 & 0.43 & 0 \tabularnewline
    \cmidrule(lr){1-4}\cmidrule(lr){5-11}
    37 & \Hard & Const. & Translation & 0.06 & 0.06 & 0.04 & 0.05 & 0.05 & 0.76 & 0 \tabularnewline
    38 & \Hard & Const. & Yaw & 0.05 & 0.06 & 0.05 & 0.05 & 0.11 & 0.74 & 0 \tabularnewline
    39 & \VeryHard & Const. & Both & 0.02 & 0.07 & 0.02 & 0.06 & 0.01 & 0.87 & 0 \tabularnewline
    \cmidrule(lr){1-4}\cmidrule(lr){5-11}
    40 & \VeryHard & Linear & Translation & 0.17 & 0.19 & 0.05 & 0.16 & 0.03 & 1.56 & 0 \tabularnewline
    41 & \VeryHard & Linear & Yaw & 0.01 & 0.12 & 0.03 & 0.13 & 1.47 & 0.72 & 0 \tabularnewline
    42 & \VeryHard & Linear & Both & 0.02 & 0.37 & 0.02 & 0.17 & 0.47 & 2.15 & 0 \tabularnewline
    \bottomrule
    \end{tabular}}
    \par\end{centering}
\end{table}

\section{Hardware Experiments Frame Alignment Pipeline Evaluation}\label{sec:hardware-experiments}

To first evaluate our sparse mapping capability, we conducted an experiment where a single agent flew at a speed of \SI{0.5}{\m/\s} for \SI{60}{\s}, tracing a \SI{2.5}{\m}-circle.
The \ac{uav} in use was equipped with both an Intel\textsuperscript{\textregistered} RealSense\textsuperscript{TM} T265 and an Intel\textsuperscript{\textregistered} NUC\textsuperscript{TM} 10. 
Note that our image segmentation-based frame alignment pipeline operated in real-time on the NUC\textsuperscript{TM} 10, powered by the Intel\textsuperscript{\textregistered} Core\textsuperscript{TM} i7-10710U Processor.
By employing FastSAM~\cite{zhao2023fast}, we produce segmented images approximately every \SI{200}{\ms}.
We arranged flat mats on the ground for this test, as visualized in Fig.~\ref*{fig:hw-multi-agent-frame-alignment-env}.

To assess our sparse mapping and segmentation-based pipeline, we utilized ground truth localization from a motion capture system.
This way, we can assess the sparse mapping capability by matching the generated data with the ground truth locations of the objects.
It also allows us to keep track of the ground truth frame discrepancy, which is zero, and, therefore, we can assess the pipeline accurately.

\begin{figure}[h]
    \centering
    \includegraphics[width=0.8\columnwidth, trim=2in 2in 4.5in 3.5in, clip]{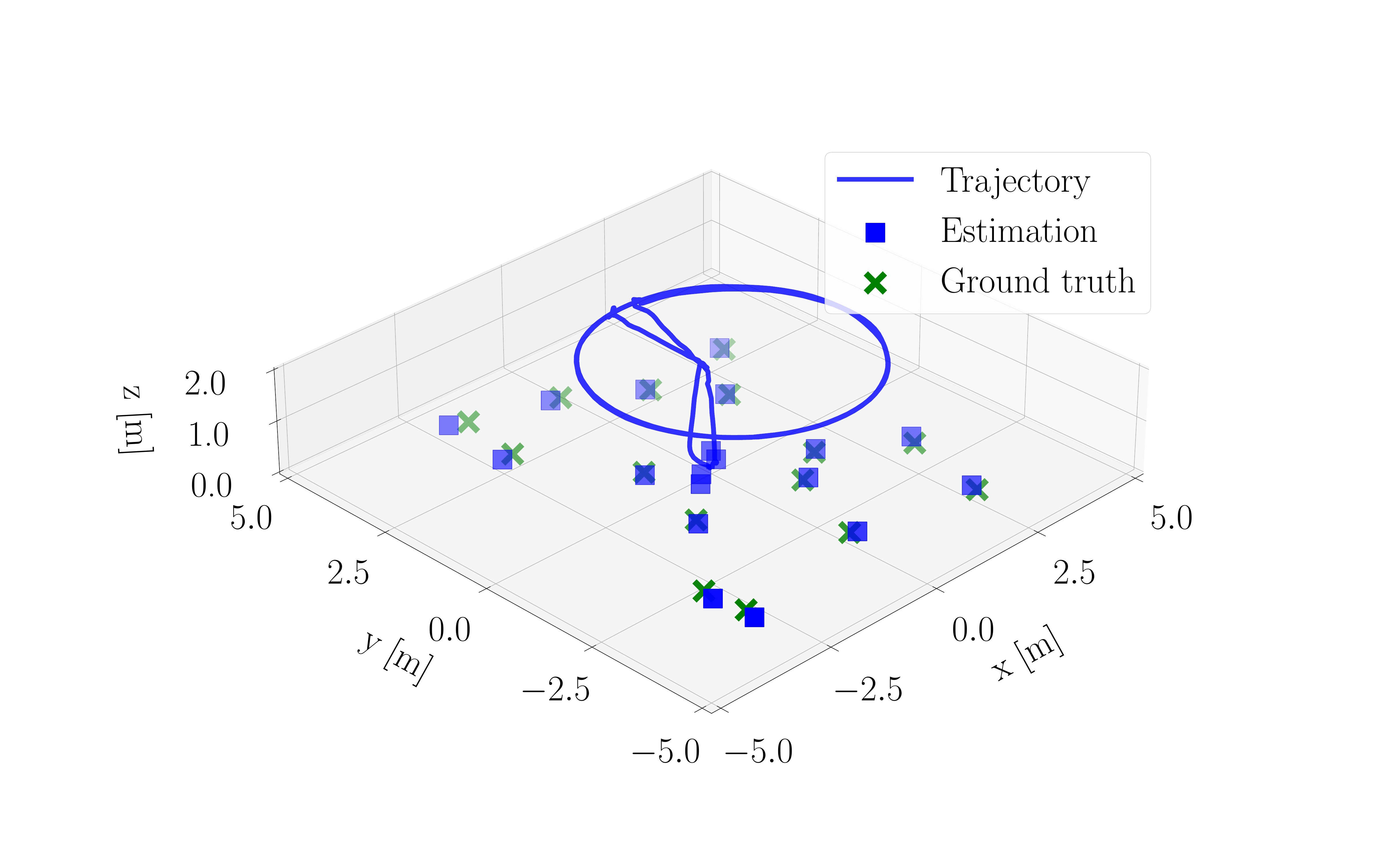}
    \caption{Single-agent hardware evaluation: agent follows \SI{2.5}{\m}-circle while successfully creating sparse map.\label{fig:hw-single-agent-sparse-mapping}}
    \vspace{-1em}
\end{figure}

The result, as illustrated in Fig.~\ref*{fig:hw-single-agent-sparse-mapping}, confirms that our image segmentation-based mapping technique effectively extracted a sparse map directly from raw images.

In addition, we performed an assessment of the effectiveness of our frame alignment pipeline in a multiagent context by having two agents navigate a \SI{2.5}{\m}-circle (see Fig.~\ref*{fig:hw-multi-agent-frame-alignment-env}).
Table~\ref*{fig:hw-multi-agent-frame-alignment-table} shows that our frame alignment method adeptly synchronized the coordinate frames of the two agents in real time.

\begin{table}[h]
    \renewcommand{\arraystretch}{1}
    \scriptsize
    \begin{centering}
    \caption{\centering Multiagent Frame Alignment Hardware Evaluation}
    \label{fig:hw-multi-agent-frame-alignment-table}
    \resizebox{1.0\columnwidth}{!}{
    \begin{tabular}{>{\centering\arraybackslash}m{0.05\columnwidth} >{\centering\arraybackslash}m{0.15\columnwidth} >{\centering\arraybackslash}m{0.18\columnwidth} >{\centering\arraybackslash}m{0.08\columnwidth} >{\centering\arraybackslash}m{0.07\columnwidth} >{\centering\arraybackslash}m{0.07\columnwidth} >{\centering\arraybackslash}m{0.07\columnwidth} >{\centering\arraybackslash}m{0.07\columnwidth} >{\centering\arraybackslash}m{0.07\columnwidth} }
        \toprule
        \multicolumn{3}{c}{\textbf{Settings}} & \multicolumn{6}{c}{\textbf{Results}} \tabularnewline
        \cmidrule(lr){1-3}
        \cmidrule(lr){4-9}
        \multirow{2}[2]{*}{\textbf{Traj.}} & \multirow{2}[2]{*}{\shortstack{\textbf{Added Offset/} \\ \textbf{Drift Type}}} & \multirow{2}[2]{*}{\shortstack{\textbf{Translation/} \\ \textbf{Yaw}}} & \multicolumn{2}{c}{\textbf{X Err. [m]}} & \multicolumn{2}{c}{\textbf{Y Err. [m]}} & \multicolumn{2}{c}{\textbf{Yaw Err. [deg]}} \tabularnewline
        &&& \textbf{Mean} & \textbf{Std.} & \textbf{Mean} & \textbf{Std.} & \textbf{Mean} & \textbf{Std.} \tabularnewline
        \cmidrule(lr){1-3}
        \cmidrule(lr){4-9}
        Circle & None & None & 0.29 & 0.36 & 0.16 & 0.26 & 2.59 & 1.77\tabularnewline
    \bottomrule
\end{tabular}}
\par\end{centering}
\end{table}

\section{Conclusions}\label{sec:conclusion}
In multiagent trajectory planning, 
the challenges of frame alignment drift and obstacle and peer collision avoidance must all be addressed.
This paper introduced a perception and uncertainty-aware planner that can navigate unknown spaces while avoiding known obstacles.
Our real-time, image segmentation-based pipeline robustly estimates frame alignment between drifted agent frames.
Future work includes larger-scale flight experiments and extending our pipeline to 3D mapping.

\balance %
\bibliographystyle{IEEEtran}
\bibliography{refs}

\end{document}